\documentclass[10pt,journal,compsoc]{IEEEtran}
\usepackage[utf8]{inputenc} 
\usepackage[T1]{fontenc}    
\usepackage{booktabs}       
\usepackage{amsfonts}       
\usepackage{subfig}

\usepackage{amsmath}
\usepackage{amssymb}

\usepackage[table]{xcolor}
\usepackage{array}
\usepackage{multirow}
\usepackage{verbatim}

\newcolumntype{C}[1]{>{\centering}p{#1}}
\newcolumntype{R}[1]{>{\raggedleft\let\newline\\\arraybackslash\hspace{0pt}}m{#1}}

\usepackage{tikz}
\usetikzlibrary{shapes.geometric}
\usetikzlibrary{shapes.arrows}
\usetikzlibrary{positioning}
\usetikzlibrary{backgrounds}
\usetikzlibrary{decorations.markings}

\tikzset{%
    square/.style = {regular polygon,
        regular polygon sides=4},
    diagonal square/.style 2 args={draw, regular polygon,
        regular polygon sides=4, fill=#2, path picture={
            \fill[#1, sharp corners] (path picture bounding box.south west) -|
            (path picture bounding box.north east) -- cycle;}},
    diagonal circle/.style 2 args={draw, circle, fill=#2, path picture={
            \fill[#1, sharp corners] (path picture bounding box.south west) -|
            (path picture bounding box.north east) -- cycle;}},
    block/.style    = {rectangle, draw, 
        font=\scriptsize,
        text width=6em, text centered,
        rounded corners, minimum height=2em 
    },
    tblock/.style = {rectangle,
        font=\scriptsize,
    },
    line/.style= {draw},
    dline/.style = {dashed},
    arrow/.style     = { draw, -> },
    darrow/.style = {dashed, ->},
}

\usepackage{pgfplots}
\pgfplotsset{compat=newest, 
    tick label style={font=\scriptsize},
    label style={font=\scriptsize},
    legend style={font=\scriptsize}}

\definecolor{brightBlue}{RGB}{68, 119, 170}
\definecolor{brightCyan}{RGB}{102, 204, 238}
\definecolor{brightGreen}{RGB}{34, 136, 51}
\definecolor{brightYellow}{RGB}{204, 187, 68}
\definecolor{brightRed}{RGB}{238, 102, 119}
\definecolor{brightPurple}{RGB}{170, 51, 119}
\definecolor{brightGrey}{RGB}{187, 187, 187}

\definecolor{vibrantBlue}{RGB}{0, 119, 187}
\definecolor{vibrantCyan}{RGB}{51, 187, 238}
\definecolor{vibrantTeal}{RGB}{0, 153, 136}
\definecolor{vibrantOrange}{RGB}{238, 119, 51}
\definecolor{vibrantRed}{RGB}{204, 51, 17}
\definecolor{vibrantMagenta}{RGB}{238, 51, 119}
\definecolor{vibrantGrey}{RGB}{100, 100, 100}

\tikzstyle{dotted}=                  [dash pattern=on \pgflinewidth off 2pt]
\tikzstyle{densely dotted}=          [dash pattern=on \pgflinewidth off 1pt]
\tikzstyle{loosely dotted}=          [dash pattern=on \pgflinewidth off 4pt]

\tikzstyle{dashed}=                  [dash pattern=on 3pt off 3pt]
\tikzstyle{densely dashed}=          [dash pattern=on 3pt off 2pt]
\tikzstyle{loosely dashed}=          [dash pattern=on 3pt off 6pt]

\tikzstyle{dashdotted}=              [dash pattern=on 3pt off 2pt on \pgflinewidth off 2pt]
\tikzstyle{densely dashdotted}=      [dash pattern=on 3pt off 1pt on \pgflinewidth off 1pt]
\tikzstyle{loosely dashdotted}=      [dash pattern=on 3pt off 4pt on \pgflinewidth off 4pt]

\tikzstyle{dash dot dot}=[dash pattern=on 3pt off 2pt on \pgflinewidth off 2pt on \pgflinewidth off 2pt]
\tikzstyle{densely dash dot dot}=[dash pattern=on 3pt off 1pt on \pgflinewidth off 1pt on \pgflinewidth off 1pt]
\tikzstyle{loosely dash dot dot}= [dash pattern=on 3pt off 4pt on \pgflinewidth off 4pt on \pgflinewidth off 4pt]

\tikzstyle{dash dash dot}=[dash pattern=on 3pt off 2pt on 3pt off 2pt on \pgflinewidth off 2pt]
\tikzstyle{densely dash dash dot}=[dash pattern=on 3pt off 1pt on 3pt off 1pt on \pgflinewidth off 1pt]
\tikzstyle{loosely dash dash dot}= [dash pattern=on 3pt off 4pt on 3pt off 4pt on \pgflinewidth off 4pt]

\tikzstyle{dash dash dot dot}=[dash pattern=on 3pt off 2pt on 3pt off 2pt on \pgflinewidth off 2pt on \pgflinewidth off 2pt]
\tikzstyle{densely dash dash dot dot}=[dash pattern=on 3pt off 1pt on 3pt off 1pt on \pgflinewidth off 1pt on \pgflinewidth off 1pt]
\tikzstyle{loosely dash dash dot dot}= [dash pattern=on 3pt off 4pt on 3pt off 4pt on \pgflinewidth off 4pt on \pgflinewidth off 4pt]

\pgfplotsset{
	cycle list/.define={vibrant}{
    	vibrantBlue, solid, every mark/.append style={solid, fill=vibrantBlue},mark=*\\
    	vibrantCyan, densely dashed, every mark/.append style={solid, fill=vibrantCyan}, mark=triangle*\\
    	vibrantTeal, densely dotted, every mark/.append style={solid, fill=vibrantTeal}, mark=square*\\
    	vibrantOrange, densely dashdotted, every mark/.append style={solid, fill=vibrantOrange}, mark=diamond*\\
        vibrantRed, densely dash dot dot, every mark/.append style={solid, fill=vibrantRed, scale=1.5}, mark=x\\
        vibrantMagenta, densely dash dash dot, every mark/.append style={solid, fill=vibrantMagenta, scale=1.5}, mark=star\\
        vibrantGrey, densely dash dash dot dot, every mark/.append style={solid, fill=vibrantGrey, scale=1.5}, mark=|\\
        },
    }

\ifCLASSOPTIONcompsoc
  \usepackage[nocompress]{cite}
\else
  \usepackage{cite}
\fi
\ifCLASSINFOpdf
\else
\fi
\begin{document}
%
\title{Distributed Deep Convolutional Neural Networks\\ for the Internet-of-Things}
%
%
%
%

\author{Simone~Disabato,
    Manuel~Roveri,~\IEEEmembership{Senior~Member,~IEEE,}
    and~Cesare~Alippi,~\IEEEmembership{Fellow,~IEEE}
    \IEEEcompsocitemizethanks{\IEEEcompsocthanksitem Simone Disabato and Manuel Roveri are with the Dipartimento di Elettronica, Informazione e Bioingegneria, Politecnico di Milano, 20133 Milano, Italy. E-mail: \{simone.disabato, manuel.roveri\}@polimi.it.
        \IEEEcompsocthanksitem Cesare Alippi is with the Dipartimento di Elettronica, Informazione e Bioingegneria, Politecnico di Milano, 20133 Milano, Italy, and also with the Università della Svizzera Italiana, 6900 Lugano, Switzerland.
        E-mail: cesare.alippi@polimi.it.}
}

%
%

\markboth{IEEE Transactions on Computers,~Vol.~70, No.~8, August~2021}%
{Disabato \MakeLowercase{\textit{et al.}}: Distributed Deep Convolutional Neural Networks for the Internet-of-Things}
%



\IEEEtitleabstractindextext{%
    \begin{abstract}
       Severe constraints on memory and computation characterizing the Internet-of-Things (IoT) units may prevent the execution of Deep Learning (DL)-based solutions, which typically demand large memory and high processing load. In order to support a real-time execution of the considered DL model at the IoT unit level, DL solutions must be designed having in mind constraints on memory and processing capability exposed by the chosen IoT technology. In this paper, we  introduce a design methodology aiming at allocating the execution of Convolutional Neural Networks (CNNs) on a distributed IoT application. Such a methodology is formalized as an optimization problem where the latency between the data-gathering phase and the subsequent decision-making one is minimized, within the given constraints on memory and processing load at the units level. 
       The methodology supports multiple sources of data as well as multiple CNNs in execution on the same IoT system allowing the design of CNN-based applications demanding autonomy, low decision-latency, and high Quality-of-Service.
    \end{abstract}
    
    }

\onecolumn
\vspace*{3cm}

{\Large IEEE Copyright Notice}

\vspace*{0.5cm}

Copyright $\copyright$ 2021 IEEE.  Personal use of this material is permitted.  Permission from IEEE must be obtained for all other uses, in any current or future media, including reprinting/republishing this material for advertising or promotional purposes, creating new collective works, for resale or redistribution to servers or lists, or reuse of any copyrighted component of this work in other works.

\vspace*{1cm}

{\Large\textbf{Distributed Deep Convolutional Neural Networks for the Internet-of-Things}}

\vspace*{0.5cm}

Simone Disabato\IEEEauthorrefmark{1}

Manuel Roveri\IEEEauthorrefmark{1}

Cesare Alippi\IEEEauthorrefmark{1}\IEEEauthorrefmark{3}

\vspace*{0.5cm}

\IEEEauthorblockA{\IEEEauthorrefmark{1}\textit{Dipartimento di Elettronica, Informazione e Bioingegneria},
    \textit{Politecnico di Milano}, Milan, Italy
}

\IEEEauthorblockA{\IEEEauthorrefmark{3} \textit{Faculty of Informatics, Universit\`{a} della Svizzera Italiana (USI)}, Lugano, Switzerland.}

\vspace*{0.5cm}

{\large Published in IEEE Transactions on Computers.}

The code of the paper is available at https://github.com/simdis/distributedCNNs.


\vfill

{\Large Please cite as:}

\vspace*{0.5cm}

\begin{minipage}[l]{0.75\textwidth}
    S. Disabato, M. Roveri and C. Alippi, "Distributed Deep Convolutional Neural Networks for the Internet-of-Things," in IEEE Transactions on Computers, vol. 70, no. 8, pp. 1239-1252, 1 Aug. 2021, doi: 10.1109/TC.2021.3062227. 
\end{minipage}

\vfill

{\Large BibTex}

\vspace*{0.5cm}

%
%
%
%
%
%
%
%

\begin{minipage}[l]{0.75\textwidth}
    
    \begin{verbatim}
        @article{disabato2021distributed,
            title={Distributed deep convolutional neural networks 
                for the internet-of-things},
            author={Disabato, Simone and Roveri, Manuel and Alippi, Cesare},
            journal={IEEE Transactions on Computers},
            year={2021},
            volume={70},
            number={8},
            pages={1239-1252},
            doi={10.1109/TC.2021.3062227},
            publisher={IEEE}
        }

    \end{verbatim}

\end{minipage}

\vfill

\clearpage

\twocolumn

\maketitle

\IEEEdisplaynontitleabstractindextext

%
\IEEEpeerreviewmaketitle

\section{Introduction}
\label{sct:introduction}
Deep learning (DL) represents the state-of-the-art in many recognition/classification applications~\cite{zhang2018deep,hof2013breaktrough,szegedy2015going,he2016deep} and is characterized by processing architectures organized in a pipeline of layers providing a hierarchical representation~\cite{lecun2015deep}. However, such solutions are also characterized by a high computational load and memory occupation~\cite{simonyan2014very,he2015convolutional,alippi2018moving} and, for this reason, their use is mostly restricted to high-performance computing platforms~\cite{dean2012large,cui2016geeps,hardy2017distributed}. 

For this reason, Internet-of-Things (IoT) systems, whose computational units are mostly constrained by limited processing, memory and energy capacities, have been rarely considered a viable technological solution for DL. In fact, currently, IoT units are generally considered as simple data-collectors acquiring and transmitting data to the Cloud for DL processing~\cite{mohammadi2018deep}, with very few solutions (mostly based on approximation techniques) proposing machine learning and (most rarely) deep learning for embedded systems~\cite{alippi2018moving,disabato2020incremental,zhao2018deepthings}. Unfortunately, the request for a remote processing to make decisions limits the effectiveness of the system as  the ``data production to decision making''-latency might not satisfy real-time constraints. The system closed-loop stability may even be compromised when remote connectivity between IoT units and Cloud is unavailable or limited in bandwidth~\cite{alippi2017not}. Hence, applications requesting a (quasi) real-time decision/actuation cannot take advantage of remote Cloud-based processing of DL solutions.  

The problem of reducing the complexity of DL solutions to match the technological constraints of IoT systems is becoming more and more relevant from both the scientific and technological perspective and solutions present in the literature address such a problem at different levels (i.e., ad-hoc hardware, approximation techniques, off-loading of DL and distributed DL).  Despite the growing research interest, the problem of assigning a distributed DL solution on a network of IoT embedded devices is still open in the literature. 

In this research direction, this paper introduces a methodology that receives technological constraints associated with (possibly heterogeneous) IoT units  and the DL trained architectures, and provides the optimal distributed assignment of the DL computation (layers) to the IoT units by minimizing the data gathering to decision latency. In particular, the methodology has been tailored to Convolutional Neural Networks (CNNs), representing the state-of-the-art in image and video processing. Interestingly, these CNNs can  operate sequentially on  the processing pipeline (all layers are executed up to the final classification)~\cite{krizhevsky2012imagenet,he2016deep,redmon2016you} or can decide the processing path  at run-time (skipping the execution of some layers) according to the information content brought by the  input~\cite{bolukbasi2017adaptive,disabato2018reducing}. Both cases are considered in the proposed methodology that, in addition, can be used with multiple CNNs running on the same network of IoT embedded devices (possibly sharing processing layers).

With respect to the literature, the main novel aspects of the proposed methodology can be summarized as:
\begin{itemize}
    \item the methodology is able to take into account communication and computation capabilities as well as memory constraints of the IoT units;
    \item the methodology can support CNNs whose processing pipeline depends on the specific input image;
    \item multiple CNNs (possibly sharing processing layers) can be executed on the same network of IoT units.
\end{itemize}
The proposed methodology has been validated both on simulations and a real-world technological scenario, while the related code is made available to the scientific community\footnote{The code is available at {\scriptsize https://github.com/simdis/distributedCNNs.}}. The paper is organized as follows: the related literature is analysed in Section~\ref{sct:literature}, the research problem is formulated in Section~\ref{sct:problemformulation}, whereas the proposed methodology is described in Section~\ref{sct:theMethodology}. Results are detailed in Sections~\ref{sct:experimentalResults} and~\ref{subsct:realImplementation}, while conclusions are drawn in Section~\ref{sct:conclusions}.
%
\vspace*{-0.25cm}
\section{Related Literature}
\label{sct:literature}
In the related literature, the problem of reducing the computational and memory demand of DL solutions to match the technological constraints of IoT systems can be addressed at different levels. 
At the hardware level, the best technological performances with minimum power consumption are achieved by ad-hoc hardware computing platforms for DL as specific processors~\cite{moons201714}, or configurable FPGA solutions~\cite{zhang2015optimizing,suda2016throughput}. As also pointed out by~\cite{cavigelli2016origami}, such solutions require high design-skills and cannot be easily re-used. Another approach, that has shown a significant drop in computation time, is the adoption of GPUs~\cite{krizhevsky2012imagenet,cui2016geeps}, TPUs~\cite{wei2019benchmarking}, or neural hardware. However, both custom hardware and GPU/TPU solutions cannot be easily considered in pervasive IoT systems.

The reduction of DL solutions complexity can be achieved by considering approximation techniques. Such an approach allows to reduce the memory and/or computational demand at the expense of a decrease in the accuracy.~\cite{han2015deep,he2017channel,lin2018holistic} proposed compression techniques on already trained CNNs, by compressing the weights, by pruning whole or part of layers, and by adopting Huffman coding. Another approach is the quantization of layers' weights~\cite{cai2017deep}, up to binary weights, also during the training phase~\cite{denton2014exploiting,courbariaux2015binaryconnect,rastegari2016xnor}. Similarly,~\cite{alippi2018moving} introduced task-dropping and precision-scaling mechanisms to design application-specific approximated CNNs able to be executed in off-the-shelf embedded systems, but not in distributed IoT systems. \cite{li2018deeprebirth}, instead, studied approximation to reduce inference on pooling and normalization layers. An approach to reduce the inference time is those of Adaptive Early Exit CNNs~\cite{bolukbasi2017adaptive} or Gate-Classification CNNs~\cite{disabato2018reducing}, where the classification output can be provided also at intermediate layers, skipping the remaining computation.

A different point of view is provided by the off-loading techniques for distributed computing systems. Here, the goal is not to reduce DL solutions' complexity but to move computationally-intensive processing to high-performing units of the distributed system. For instance, a framework to optimally offload code in a pervasive system comprising mobile units is proposed in~\cite{shi2012serendipity}, by either minimizing the total communication latency or the overall energy consumption. Differently,~\cite{hong2013mobile} proposed a high-level programming language to design applications meant for Fog-Computing Sensor Networks able to hide the heterogeneity of computing nodes and their position in space.~\cite{hu2019throughput} proposed a low-complexity scheduler that increases the throughput in IoT clusters by relying on tasks duplication and splitting by taking into account communication and computation capabilities, but not IoT units memory constraints.
Very few works present in the literature encompass the code-offloading of machine learning-based applications in pervasive systems, e.g.,~\cite{cheng2015just} where the classification/pattern recognition tasks of a wearable device are partially offloaded to other computing units (e.g., mobile phones). Similarly to our vision, here the priority in the offloading is to move code at first to the closest mobile devices and, then, if needed, to the Cloud.

The problem of distributing DL solutions has been recently addressed in the field of edge and fog computing.~\cite{teerapittayanon2017distributed} introduced a distributed framework for CNNs operating in edge computing platforms, with the possibility of distributing the CNNs computation, mostly restricted to the Cloud, also to edge or local devices. Unfortunately, the usage of the Cloud is predominant in this work.
In~\cite{alippi2017not,teerapittayanon2017distributed} emerged the need to completely re-design the DL solutions to take into account hardware and physical constraints at application design-time, to make IoT systems a viable technological solution for DL. Such an approach is considered in some recent works addressing the problem of distributing machine/deep learning solutions onto a  network of IoT/Edge devices~\cite{bhardwaj2019memory,chen2019distributed,hu2020fast,tao2018esgd,zhao2018deepthings}. In particular,~\cite{zhao2018deepthings} organizes the computation tasks by partitioning the convolutional layers vertically, with the possibility of sharing the computed features among parallel tasks.~\cite{hu2020fast}, instead, proposed a dynamic scheduler to assign the layers to edge units and reduce the communication data by relying on autoencoders.

Unfortunately, almost all these solutions rely on approximations of the original ML or DL solution to match the hardware and physical constraints of the devices. In addition, the processing pipelines of the machine/deep learning solutions introduced in these works are fixed at design-time, hence not being able to modify their execution at run-time according to the information content brought by the input. Finally, the proposed solutions are not meant to be executed on low-power embedded systems (e.g., STM32 or Arduino).
%
\section{Problem Formulation}
\label{sct:problemformulation}
The IoT system comprises a set of $C$ data-acquisition units $\{s_1, \ldots, s_C\}$ mounting cameras and providing the images to be classified by the $C$ CNNs (each CNN processes data coming from one data-acquisition unit), a set $\mathbb{N}_N = \{1, \dots, N\}$ of $N$ possibly heterogeneous IoT units implementing code execution, and one target unit $f$ receiving all the $C$ CNNs' outcomes to make a decision or activate a reaction. Without any loss of generality, the $C$ data-acquisition units are assumed to only acquire images and do not participate in the computation.
The $i$-th IoT unit $i \in \mathbb{N}_N$ is characterized by its own constraints w.r.t. maximum memory capacity $\bar{m}_i$ and available computation $\bar{c_i}$. 

The IoT system is modeled as a graph $G(V,E)$ of nodes $V= \{\mathbb{N}_N \cup \left\{s_1,\dots,s_C,f\right\}\}$ and arcs $E$. An arc $e_{i_1,i_2}$ between unit $i_1$ and $i_2$ exists in $E$ if $i_2$ is within the range of the transmission technology the IoT unit $i_1$ is equipped with\footnote{All the units in $V$ are assumed to share the same transmission technology with a fixed transmission data-rate. If the IoT units $i_1$ and $i_2$ adopt two different transmission technologies such that $i_2$ is within the transmission range of $i_1$, but $i_1$ is not inside the one of $i_2$, then $d_{i_1,i_2} = 1 \not= d_{i_2,i_1}$ (loss of symmetry property).}. 
Let $d_{i_1,i_2}$, for each $i_1,i_2 \in V$, be the \textit{hop-distance} between units $i_1$ and $i_2$ of $V$ defined as the number of hops (communication steps), data need to take to reach $i_2$ from $i_1$. In other terms, distance $d_{i_1,i_2}$ is the shortest communication path between units $i_1$ and $i_2$ within the graph $G$. Following the definition of shortest path in a graph, if no path between unit $i_1$ and $i_2$ exists, then $d_{i_1,i_2}=+\infty$.  We also assume that no isolated node exists, i.e., $d_{i_1,i_2} < \infty$, for each $i_1,i_2 \in V$.

Let $M_u$, for each $u \in \mathbb{N}_C= \{1, \ldots, C\}$, be the number of layers characterizing the $u$-th CNN. 
Each layer $j$ of CNN $u$, for each $j \in \{ 1,\ldots,M_u \} $ and $u \in \mathbb{N}_C$, is characterized by a given memory demand $m_{u,j}$ and computation $c_{u,j}$. More specifically, the memory complexity $m_{u,j}$ (in Bytes) is defined as the number of weights that layer $j$ of CNN $u$ has to store multiplied by the size of the data type used to represent those parameters (typically the floating-point type occupying 4 Bytes), while the computational load $c_{u,j}$ is measured as the number of multiplications to be executed by that layer~\cite{alippi2018moving}. 
Let $K_{u,j}$, for each $u \in \mathbb{N}_C$ and $j \in \{ 1,\ldots,M_u \}$, be the memory occupation of the intermediate representation transmitted from layer $j$ to the subsequent layer $j+1$ of CNN $u$, and $K_{u,s}$ and $K_{u,M_u}$ be the memory occupation of the input image of CNN $u$ transmitted from source $s_u$ to the unit executing the first layer of the $u$-th CNN and the final classification provided by layer $M_u$ sent to the target unit $f$, respectively. In particular, $K_{u,M_u}$ is either the classification label or the posterior probability of the classes resulting from the softmax layer. 

When the processing path of a CNN depends on the information content, such as in Adaptive~\cite{bolukbasi2017adaptive} or Ga\-te-Clas\-si\-fi\-ca\-tion CNNs~\cite{disabato2018reducing}, classification completes as soon as enough confidence is achieved (the execution of the remaining layers is aborted).
To achieve this goal, Early-Exit CNNs (EX-CNNs) are endowed with intermediate exit points, creating multiple paths within the CNNs each of which is characterized by a probability of being traversed. More formally, given a $M_u$-layer EX-CNN, let $p_{u,j} \in [0,1]$ be the probability that the $j$-th layer of the $u$-th CNN processes the input image\footnote{The probabilities $p_{u,j}$s can be estimated during CNN learning.}   and let $g_{u,j} \in [0,1]$, for each $u \in \mathbb{N}_C$ and $j \in \{1,\ldots,M_u\}$, be the probability that the computation ends at layer $j$ of the $u$-th CNN is
\begin{equation}
    g_{u,j}= 
    \begin{cases}
        p_{u,j} - p_{u,j+1}  &\text{ if }  j < M_u  \\
        1 - \sum_{v=1}^{M_u-1} g_{u,v} &\text{ if }   j=M_u
        %
    \end{cases}.
    \label{eq:gjgateclassifiertosink}
\end{equation}
The addressed problem is the optimal placement of the $C$ CNNs layers on the $N$ IoT units to minimize the latency in transmitting decisions about the images gathered by the $C$ sources to the target unit $f$.
%
\section{The Proposed Methodology}
\label{sct:theMethodology}
This section introduces the proposed methodology for the optimal placement of  CNNs on the IoT system. Such a methodology has been reformulated as a decision-making latency minimization optimization problem aiming at optimally assigning the layers of the $C$ CNNs to the $N$ IoT units.
This optimization problem relies on the $C$$N$$M$ variables $\alpha_{u,i,j}$ defined as:
\begin{equation}
    \alpha_{u,i,j} =
    \begin{cases}
        1  &\text{\small if IoT unit $i$ executes layer $j$ of CNN $u$} \\
        0 &\text{\small otherwise}
    \end{cases},
    \label{eq:variablealphacij}
\end{equation}
for each $u \in \mathbb{N}_C$, for each $i \in \mathbb{N}_N$ and for each $j \in \mathbb{N}_M = \{1, \ldots, M\}$, being $M=max \{ M_1,\ldots,M_u \}$ the maximum number of layers among the considered $C$ CNNs (i.e., the maximum depth of all CNNs).

Without loss of generality, the distances $d_{i_1,i_2}$, for each $i_1,i_2 \in V$, can be precomputed, allowing us to define an integer quadratic optimization problem on variables $\alpha_{u,i,j}$s. As detailed in Section~\ref{sct:problemformulation}, $\{s_1,\ldots,s_C\}$ and $f$ do not participate in the optimization problem since their task is to acquire images and receive the final classification, respectively. This assumption can be easily removed by considering additional IoT computing units in the same positions of $s_i$s and $f$.

The objective function to be minimized models the latency in making a decision by the $C$ CNNs placed on the IoT units, defined as the time occurring between images acquisition by sensor  unit $s_u$s (size $K_{u,s}$) and final classifications $K_{u,M_u}$s are transmitted to unit $f$:
\begin{multline}
    \sum_{u=1}^{C}\sum_{i=1}^{N}\sum_{k=1}^{N}\sum_{j=1}^{M-1} \alpha_{u,i,j}\cdot\alpha_{u,k,j+1}\cdot p_{u,j+1} \cdot \frac{K_{u,j}}{\rho}\cdot d_{i,k} \\
    + \sum_{i=1}^{N} t_{i}^{(p)} + t_{s} + t_{f},
    \label{eq:objectivefunction}
\end{multline}
such that
\begin{align}
    &\forall i \in \mathbb{N}_N&&\sum_{u=1}^{C}\sum_{j=1}^{M} \alpha_{u,i,j} \le L 
    \label{eq:nodeconstraint} \\
    &\forall i \in \mathbb{N}_N&&\sum_{u=1}^{C}\sum_{j=1}^{M} \alpha_{u,i,j} \cdot m_{u,j} \le \bar{m}_i \label{eq:memoryconstraint} \\
    &\forall i \in \mathbb{N}_N&&\sum_{u=1}^{C}\sum_{j=1}^{M} \alpha_{u,i,j} \cdot c_{u,j} \le \bar{c}_i \label{eq:computationconstraint} \\
    &\forall u \in \mathbb{N}_C, \forall j \in \mathbb{N}_M&&\sum_{i=1}^{N} \alpha_{u,i,j} = 
    \begin{cases}
        1 &\text{if } j \le M_u \\
        0 &\text{otherwise}
    \end{cases} 
    \label{eq:assignedlayerconstraint}
\end{align}
and where
\begin{alignat}{2}
    &t_{s} &&= \sum_{u=1}^{C}\sum_{i=1}^{N} \alpha_{u, i,1} \cdot p_{u,1} \cdot \frac{K_{u,s}}{\rho}\cdot d_{s,i} 
    \label{eq:sourcetime} \\
    &t_{i}^{(p)} &&= \sum_{u=1}^{C}\sum_{j=1}^{M} \alpha_{u,i,j} \cdot p_{u,j}\cdot \frac{c_{u, j}}{e_i}
    \label{eq:processingtime} \\
    &t_{f} &&= \sum_{u=1}^{C}\sum_{i=1}^{N}\sum_{j=1}^{M} \alpha_{u, i, j} \cdot g_{u,j}  \cdot \frac{K_{u,M_u}}{\rho}\cdot d_{i,f}  
    \label{eq:sinktime} 
\end{alignat}
The objective function in Eq.~\eqref{eq:objectivefunction} comprises four different components of the latency:

(i) The source time $t_{s}$, defined in Eq.~\eqref{eq:sourcetime}, required to transmit the images from the sources $s_u$s to the IoT units executing the first layer of the CNNs. Although the first layer is always reached, i.e., $p_{u,1}=1$ for each $u \in \mathbb{N}_C$, the term $p_{u,1}$ has been added to Eq.~\eqref{eq:sourcetime} to provide homogeneity in the formalization. 

(ii) The transmission time of  intermediate representations among the IoT units processing the CNN layers. More precisely, the transmission time of the intermediate representation of the $j$-th layer of the $u$-th CNN from unit $i$ to $k$ is
\begin{equation}
    \frac{K_{u,j}}{\rho} \cdot d_{i,k},
    \label{eq:transmissiontime}
\end{equation}
where $\rho$ is the data-rate of the considered transmission technology and $d_{i,k}$ is the hop-distance between unit $i$ and $k$ as defined in Section~\ref{sct:problemformulation}. In Eq.~\eqref{eq:objectivefunction} the transmission time is weighted by the probability $p_{u,j+1}$ that layer $j+1$ is executed right after layer $j$.

(iii) The processing time $t_{i}^{(p)}$ of the CNN layers on the IoT units. Specifically, the processing time of  layer $j$ of  CNN $u$ on the $i$-th IoT unit is approximated as the ratio between the computational demand $c_{u,j}$ that layer requires and the number of multiplications $e_i$ the IoT unit $i$ is able to carry out in one second\footnote{The $e_i$s encompass the number of available cores, the type of pipeline such cores implement to approach one operation per clock cycle, the presence or not of a GPU allowing to parallelize CNN operations (e.g., the convolutions) and all the delays resulting from the processing system and memory management.}. In Eq.~\eqref{eq:processingtime}, the processing time is weighted by the probability $p_{u,j}$ that the layer $j$ of CNN $u$ is executed.

(iv) The sink time $t_{f}$ required to transmit the final classification $K_{u,M_u}$, for each $u \in \mathbb{N}_C$,  from the IoT units taking these decisions to the target unit $f$. It is noteworthy to point out that Eq.~\eqref{eq:sinktime} takes into account all feasible output paths from node $i$ to the target unit $f$, suitably weighted by the probability $g_{u,j}$ that the classification is made at layer $j$ of CNN $u$ in execution on IoT unit $i$.

The constraint in Eq.~\eqref{eq:nodeconstraint} ensures that each IoT unit contains at most $L$ layers, being $L$ an additional user-defined model hyper-parameter. In particular, when $L=1$, at most one layer can be assigned to an IoT unit, while $L>1$ implies that up to $L$ layers (also belonging to different CNNs) can be assigned to a particular IoT unit. The constraints in Eq.~\eqref{eq:memoryconstraint} and~\eqref{eq:computationconstraint} are meant to take into account the technological limits about memory usage and computational load characterizing each IoT unit. Finally, the constraint in Eq.~\eqref{eq:assignedlayerconstraint} ensures that each layer $j$, for each $j \in \mathbb{N}_M$, is assigned to exactly one node and, at the same time, manages the possibility that the $C$ CNNs might be characterized by a different number $M_u \le M$ of layers, for each $u \in \mathbb{N}_C$. In fact, in those cases (i.e., when $M_u < M$), the unneeded $\alpha_{u,i,j}$s are set to $0$.

When the $j_1$-th layer of CNN $u_1$ and the $j_2$-th layer of CNN $u_2$ are shared between the two CNNs, the following constraint is added to the optimization problem
\begin{equation}
    \forall i \in \mathbb{N}_N \qquad \alpha_{u_1,i,j_1} = \alpha_{u_2,i,j_2},
    \label{eq:sharedlayerequalplacemet}
\end{equation}
to ensure that the shared layer is placed on the same IoT unit. In addition, the constraints on the maximum number of layers placed on a IoT unit - Eq.~\eqref{eq:nodeconstraint} - and the memory usage and computational load constraints - Eqs~\eqref{eq:memoryconstraint} and~\eqref{eq:computationconstraint} - are modified as follows to count the shared layer only once:
\begin{align}
    &\forall i \in \mathbb{N}_N&&\sum_{u=1}^{C}\sum_{j=1}^{M} \alpha_{u,i,j} 
    \le 
    L + \alpha_{u_2,i,j_2} ,
    \label{eq:sharedlayrenodeconstraint} \\
    &\forall i \in \mathbb{N}_N&&\sum_{u=1}^{C}\sum_{j=1}^{M} \alpha_{u,i,j} \cdot m_{u,j} 
    \le 
    \bar{m}_i + \alpha_{u_2,i,j_2}\cdot m_{u_2,j_2},
    \label{eq:sharedlayermemoryconstraint} \\
    &\forall i \in \mathbb{N}_N&&\sum_{u=1}^{C}\sum_{j=1}^{M} \alpha_{u,i,j} \cdot c_{u,j} 
    \le 
    \bar{c}_i + \alpha_{u_2,i,j_2}\cdot c_{u_2,j_2}.
    \label{eq:sharedlayercomputationconstraint}
\end{align}
If a layer is shared among $k$ CNNs, the Eqs.~\eqref{eq:sharedlayrenodeconstraint},~\eqref{eq:sharedlayermemoryconstraint}, and~\eqref{eq:sharedlayercomputationconstraint} need to take into account $k-1$ out of the $k$ variables corresponding to the shared layer.

The considered class of optimization problems, i.e., the integer quadratic programs, is NP-complete. More specifically, since the $\alpha_{u, i, j}$s are binary variables, it is possible to convert it to a binary linear program, which is one of Karp's 21 NP-complete problems~\cite{karp1972reducibility}. In the proposed methodology, the optimization problem is solved through the Gurobi solver\footnote{The solver is able to find a solution with a error less than 2\% with respect to the optimal placement in less than 2 seconds and a negligible memory occupation in all the considered IoT scenarios described in Section 5.}.

The optimization problem outcome is the optimal placement $\alpha_{u,i,j}$s of the $C$ CNNs' layers to the $N$ IoT units minimizing the delay in making a classification.
In the event that the optimization provides more than one solution, the optimal placement is any feasible solution with minimal latency. 
In the rest of this section, the methodology is tailored to three specific configurations of the DL solution, followed by some comments about the open points.
\subsection{The configuration with a single CNN}
\label{subsct:singleCNN}
\begin{figure*}[!t]
	\centering
	%
	%
	\subfloat[][The architecture of the considered 5-layer CNN, where $\mathcal{I}$ is the input image.]
	{
            
			\begin{tikzpicture} [node distance = 0.05 cm,align=center]%
				\centering
				\node (input) [text centered] { $\mathcal{I}$};
				\node (conv1) [block, right = 0.4cm of input] {  (L1) 5x5 conv, 32 \\ 2x2 max-pool, /2};
				\node (conv2) [block, right = 0.4cm of conv1] { (L2) 5x5 conv, 64 \\ 2x2 max-pool, /2 };
				\node (fc3) [block, right = 0.4cm of conv2] { (L3) FC 384};	
				\node (fc4) [block, right = 0.4cm of fc3] { (L4) FC 192};
				\node (fc5) [block, right = 0.4cm of fc4] { (L5) FC 10};
				\node (output) [circle, fill=black,minimum size=0.2cm, inner sep=0pt,right = 0.5cm of fc5] {};	
				
				\draw [->, draw] (input) -- (conv1);
				\draw [->, draw] (conv1) -- (conv2);
				\draw [->, draw] (conv2) -- (fc3);
				\draw [->, draw] (fc3) -- (fc4);
				\draw [->, draw] (fc4) -- (fc5);
				\draw [->, draw] (fc5) -- (output);
				\label{fig:5layercnn}
			\end{tikzpicture}
	}
	
	\vspace*{-0.3cm}
	%
	\subfloat[][An example of IoT system with STM32H7 (circles) and Raspberry 3B+ (squares). The source $s$ and the sink $f$ share the same IoT unit. The dotted circle refers to the transmission range, equal for all the IoT units.]
	{
		\begin{tikzpicture} [node distance = 0.05 cm, auto,align=center,]%
		\centering
        \clip (-2.9,-3.1) rectangle (5.2, 1.85);
		%

		\node at (0,0) [block, text width=3em,draw=green!50!black,fill=green!50!white] (source) {$s,f$};
		\node at (2.75,-1.6) [circle,draw] (n1) {\scriptsize n\textsubscript{01}};
		%
		\node at (4.25,-2.5) [circle,draw] (n2) {\scriptsize n\textsubscript{02}};
		\node at (3.,0.8) [square,draw] (n3) {\scriptsize n\textsubscript{03}};
		\node at (0.95,-2.3) [square,draw] (n4) {\scriptsize n\textsubscript{04}};
		\node at (-0.85,-1.2) [circle,draw] (n5) {\scriptsize n\textsubscript{05}};
		\node at (1.65,-0.3) [circle,draw] (n6) {\scriptsize n\textsubscript{06}};
		\node at (-0.7,0.75) [circle,draw] (n7) {\scriptsize n\textsubscript{07}};
		\node at (0.9,1.45) [circle,draw] (n8) {\scriptsize n\textsubscript{08}};
		\node at (-2.2,-0.65) [circle,draw] (n9) {\scriptsize n\textsubscript{09}};
		\node at (-1.3,-2.6) [square,draw] (n10) {\scriptsize n\textsubscript{10}};
		\node at (4,-0.65) [square,draw] (n11) {\scriptsize n\textsubscript{11}};
		%
		\draw (2.125,0) [dashdotted, color=green!50!black, radius=2.125cm, start angle=0, end angle=180] arc;
		\draw (2.125,0) [dashdotted, color=green!50!black, radius=2.125cm, start angle=0, end angle=-180] arc;
		\label{fig:pervasivesystem}
		\end{tikzpicture}
	}
	\hfill
	\subfloat[][An example of the methodology outcome where the layers L1,\textellipsis, L5 of the 5-layer CNN in Figure~\ref{fig:5layercnn} are placed on the IoT units of the system shown in Figure~\ref{fig:pervasivesystem}, when setting $L=1$.]
	{
        \begin{tikzpicture} [node distance = 0.05 cm]%
					\centering
            \clip (-2.9,-3.1) rectangle (5.2, 1.85);
            %

            \node at (0,0) [block, text width=3em,draw=green!50!black,fill=green!50!white] (source) {$s,f$};
            \node at (2.75,-1.6) [circle,draw,fill=blue!30!white] (n1) {\scriptsize n\textsubscript{01}};
            %
            \node at (4.25,-2.5) [circle,draw] (n2) {\scriptsize n\textsubscript{02}};
            \node at (3.,0.8) [square,draw] (n3) {\scriptsize n\textsubscript{03}};
            \node at (0.95,-2.3) [square,draw,fill=blue!30!white] (n4) {\scriptsize n\textsubscript{04}};
            \node at (-0.85,-1.2) [circle,draw,fill=blue!30!white] (n5) {\scriptsize n\textsubscript{05}};
            \node at (1.65,-0.3) [circle,draw,fill=blue!30!white] (n6) {\scriptsize n\textsubscript{06}};
            \node at (-0.7,0.75) [circle,draw] (n7) {\scriptsize n\textsubscript{07}};
            \node at (0.9,1.45) [circle,draw] (n8) {\scriptsize n\textsubscript{08}};
            \node at (-2.2,-0.65) [circle,draw] (n9) {\scriptsize n\textsubscript{09}};
            \node at (-1.3,-2.6) [square,draw,fill=blue!30!white] (n10) {\scriptsize n\textsubscript{10}};
            \node at (4,-0.65) [square,draw] (n11) {\scriptsize n\textsubscript{11}};
            \node [text centered, right= 0.0cm of n5] {\footnotesize \textcolor{blue}{L1}};
            \node [text centered, above right= -0.3 and 0.0cm of n10] {\footnotesize \textcolor{blue}{L2}};
            \node [text centered, right= 0.0cm of n4] {\footnotesize \textcolor{blue}{L3}};
            \node [text centered, right= 0.0cm of n1] {\footnotesize\textcolor{blue}{L4}};
            \node [text centered, right= 0.1cm of n6] {\footnotesize\textcolor{blue}{L5}};
            %
            \draw (2.125,0) [dashdotted, color=green!50!black, radius=2.125cm, start angle=0, end angle=180] arc;
            \draw (2.125,0) [dashdotted, color=green!50!black, radius=2.125cm, start angle=0, end angle=-180] arc;

			\draw [blue, ->] (source) -- (n5);
			\draw [blue, ->] (n5) -- (n10);
			\draw [blue, ->] (n10) -- (n4);
			\draw [blue, ->] (n4) -- (n1);
			\draw [blue, ->] (n1) -- (n6);
			\draw [blue, ->] (n6) -- (source.0);
	 		\label{fig:singleCNNplacement}
		\end{tikzpicture}
	}

	\vspace*{-0.4cm}
	\subfloat[][The variables $\alpha_{u,i,j}$s representing the methodology outcome for the solution shown in Figure~\ref{fig:singleCNNplacement}, with $u=1$.]{
		\scriptsize
		\begin{tabular}{C{0.75cm}C{0.75cm}C{0.75cm}C{0.75cm}C{0.75cm}C{0.75cm}C{0.75cm}C{0.75cm}C{0.75cm}C{0.75cm}C{0.75cm}C{0.75cm}@{}m{0pt}@{}}
			&n\textsubscript{01}&n\textsubscript{02}&n\textsubscript{03}&n\textsubscript{04}&n\textsubscript{05}&n\textsubscript{06}&n\textsubscript{07}&n\textsubscript{08}&n\textsubscript{09}&n\textsubscript{10}&n\textsubscript{11}& \\ \midrule[1.25pt]
			L1&0&0&0&0&\cellcolor{blue!30!white}\textbf{1}&0&0&0&0&0&0& \\ 
			L2&0&0&0&0&0&0&0&0&0&\cellcolor{blue!30!white}\textbf{1}&0& \\ 
			L3&0&0&0&\cellcolor{blue!30!white}\textbf{1}&0&0&0&0&0&0&0& \\ 
			L4&\cellcolor{blue!30!white}\textbf{1}&0&0&0&0&0&0&0&0&0&0& \\ 
			L5&0&0&0&0&0&\cellcolor{blue!30!white}\textbf{1}&0&0&0&0&0& \\ \bottomrule[1.25pt]
		\end{tabular}
		\label{fig:tabalphaij}
	}
	\caption{The methodology is applied to a 5-layer CNN (i.e., $C=1$).}
	\label{fig:singlecnnexample}
	%
\end{figure*}
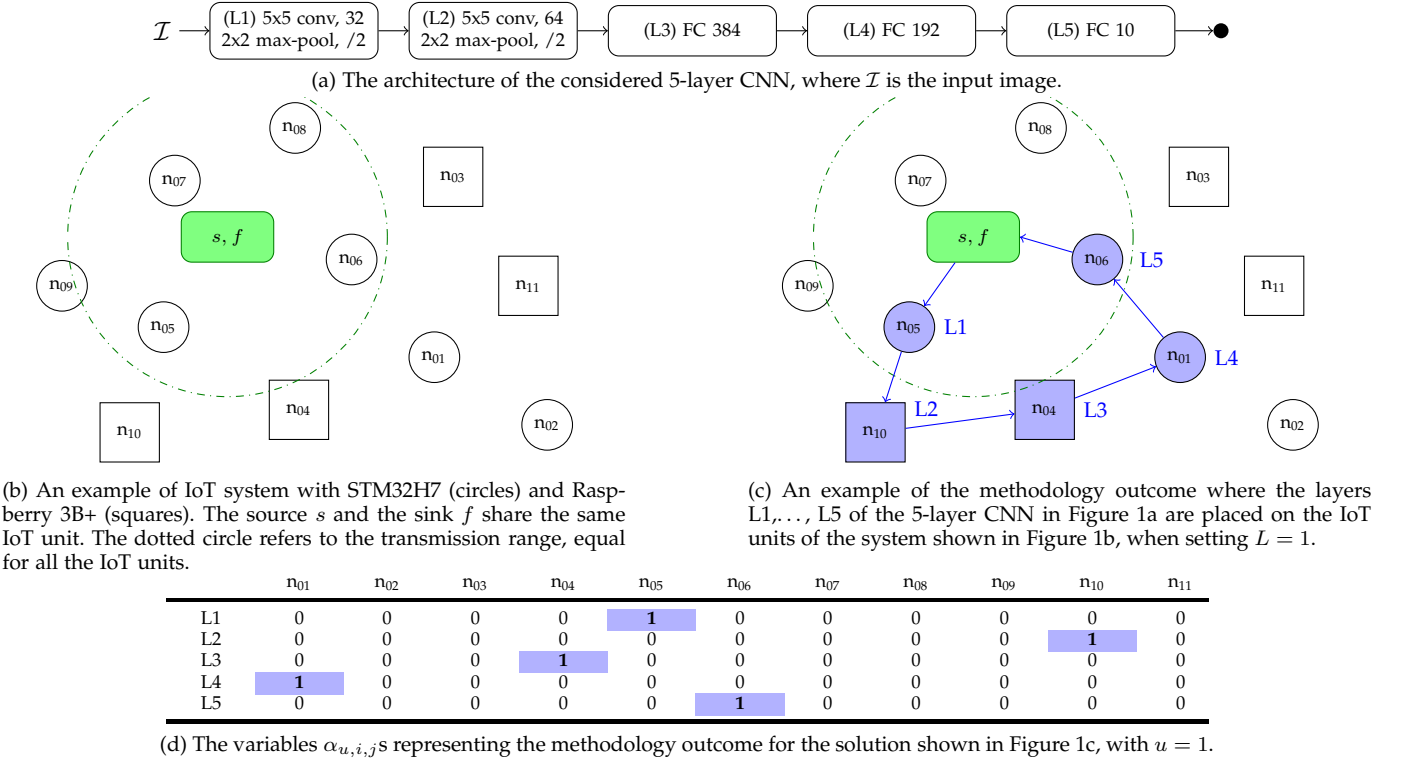
The presence of a single CNN ($C=1$) allows us to rely on $N$$M$ binary variables $\alpha_{i,j}$, determining whether layer $j$ of the CNN is assigned to unit $i$ of the IoT system.
\begin{equation}
    \alpha_{i,j} =
    \begin{cases}
        1  &\text{ if IoT unit $i$ executes CNN layer $j$} \\
        0 &\text{ otherwise}
    \end{cases},
    \label{eq:variablealphaij}
\end{equation}
for each $i \in \mathbb{N}_N$ and $j \in \mathbb{N}_M$. The objective function in Eq.~\eqref{eq:objectivefunction} modelling the latency in making a decision to be minimized is reformulated as:
\begin{equation}
    \sum_{i=1}^{N}\sum_{k=i}^{N}\sum_{j=1}^{M-1} \alpha_{i,j}\cdot\alpha_{k,j+1}\cdot \frac{K_j}{\rho} \cdot d_{i,k} + \sum_{i=1}^{N} t_{i}^{(p)} + t_s + t_f.
    \label{eq:singlecnnobjectivefunction}
\end{equation}
Then, the constraints in Eqs.~\eqref{eq:nodeconstraint},~\eqref{eq:memoryconstraint},~\eqref{eq:computationconstraint} and~\eqref{eq:assignedlayerconstraint} become:
\begin{align}
    &\forall i \in \mathbb{N}_N &&\sum_{j=1}^{M} \alpha_{i,j} \le L,
    \label{eq:singlecnnnodeconstraint} \\
    &\forall i \in \mathbb{N}_N &&\sum_{j=1}^{M} \alpha_{i,j} \cdot m_j \le \bar{m}_i,
    \label{eq:singlecnnmemoryconstraint} \\
    &\forall i \in \mathbb{N}_N &&\sum_{j=1}^{M} \alpha_{i,j} \cdot c_j \le \bar{c}_i,
    \label{eq:singlecnncomputationconstraint} \\
    &\forall j \in \mathbb{N}_M &&\sum_{i=1}^{N} \alpha_{i,j} = 1,
    \label{eq:singlecnnassignedlayerconstraint}
\end{align}
while
\begin{alignat}{2}
    &t_s &&= \sum_{i=1}^{N} \alpha_{i,1} \cdot \frac{K_s}{\rho} \cdot d_{s,i},
    \label{eq:singlecnnsourcetime} \\
    &t_{i}^{(p)} &&= \sum_{j=1}^{M} \alpha_{i,j} \cdot \frac{c_{j}}{e_i},
    \label{eq:singlecnnprocessingtime} \\
    &t_f &&= \sum_{i=1}^{N} \alpha_{i,M} \cdot \frac{K_M}{\rho} \cdot d_{i,f},
    \label{eq:singlecnnsinktime}
\end{alignat}
account for the transmission time between the source $s$ and the IoT unit running the first layer of the CNN, the processing time on all unit $i$s, and the transmission time between the unit running the $M$-th layer of the CNN and the sink $f$, respectively. 

In this configuration, the methodology has been applied to the 5-layer CNN described in Figure~\ref{fig:5layercnn}, characterized by $M=5$ layers and whose details are in Table~\ref{tab:6layergateclassification}. The considered IoT system is the one described in Figure~\ref{fig:pervasivesystem} comprising $N=11$ IoT units and being $s$ and $f$ the same unit. The IoT units belong to two different technological families, i.e., STM32H7 (round nodes) and Raspberry Pi 3B+ (squared nodes), whose memory $\bar{m}_i$ and computational $\bar{c}_i$ constraints are detailed in Table~\ref{tab:nodes}.
An example of the optimization problem outcome in this technological scenario with $L=1$ is given in Figure~\ref{fig:singleCNNplacement}, whose corresponding $\alpha_{i,j}$s are detailed in Figure~\ref{fig:tabalphaij}. 
In the optimal placement, involving three STM32H7 units (nodes $n_{05}$, $n_{01}$, and $n_{06}$) and two Raspberry Pi 3B+ (nodes $n_{10}$, and $n_{04}$), the layer $L3$ of the CNN has been assigned to a Raspberry Pi 3B+ IoT unit (i.e., $n_{04}$) since its execution on STM32H7 would violate the memory constraint.
\subsection{The configuration with a single early-exit CNN}
\label{subsct:singlegateclassificationCNN}
This configuration refers to the case where a single EX-CNN ($C=1$) has to be placed on the IoT system.  
Here, the $p_{u,j}$s and $g_{u,j}$s are simplified as $p_{j}$ and $g_{j}$, for each $j \in \mathbb{N}_M$, defining the probabilities that layer $j$ is executed and that the final classification is made at layer $j$ (i.e., the direct path from $j$ to the sink is traversed), respectively. 

More specifically, the problem variables are simplified into $N$$M$ binary variables $\alpha_{i,j}$ defined in Eq.~\eqref{eq:variablealphaij}.
Instead, the $p_{u,j}$s and $g_{u,j}$s are simplified as $p_{j}$ and $g_{j}$, for each $j \in \mathbb{N}_M$, as detailed in Section~\ref{subsct:singlegateclassificationCNN}.

The objective function modelling the latency in decision making defined in Eq~\eqref{eq:objectivefunction} is tailored as follow:
\begin{equation}
    \sum_{i=1}^{N}\sum_{k=i}^{N}\sum_{j=1}^{M-1} \alpha_{i,j}\cdot\alpha_{k,j+1}\cdot p_{j+1}\cdot \frac{K_j}{\rho} \cdot d_{i,k}
    + \sum_{i=1}^{N} t_i^{(p)} + t_s + t_f,
    \label{eq:singlegateclassificationcnnobjectivefunction}
\end{equation}
with constraints as in Eqs.~\eqref{eq:singlecnnnodeconstraint},~\eqref{eq:singlecnnmemoryconstraint},~\eqref{eq:singlecnncomputationconstraint}, and~\eqref{eq:singlecnnassignedlayerconstraint}, and where the source time $t_s$, the processing time $t_ i^{(p)}$, and the sink time $t_f$ have been modified as follows:
\begin{alignat}{2}
    &t_s &&= \sum_{i=1}^{N} \alpha_{i,1} \cdot p_1 \cdot \frac{K_s}{\rho} \cdot d_{s,i},
    \label{eq:singlegateclassificationcnnsourcetime} \\
    &t_{i}^{(p)} &&= \sum_{j=1}^{M} \alpha_{i,j} \cdot p_{j}\cdot \frac{c_{j}}{e_i},
    \label{eq:singlegateclassificationcnnsinglecnnprocessingtime} \\
    &t_f &&= \sum_{i=1}^{N}\sum_{j=1}^{M} \alpha_{i,j} \cdot g_j \cdot \frac{K_M}{\rho} \cdot d_{i,f}. \label{eq:singlegateclassificationcnnsinktime}
\end{alignat}

A 6-layer EX-CNN is shown, as an example, in Figure~\ref{fig:6layergateclassificationcnn} and detailed in Section~\ref{subsct:cnn}--Table~\ref{tab:6layergateclassification}, where $M=6$ and the Early Exit is at layer $j=2$, with a probability $\nu=0.99$ of taking the final classification. Hence, $p_1=p_2=1$ and $p_3=p_4=p_5=0.01$, with the $g_j$s different from zero only at the Early-Exit ($g_2=0.99$) and the last layer ($g_6=0.01$).
In Figure~\ref{fig:singlegateclassificationcnnexample}, the proposed methodology is applied to this CNN and the IoT system already described in Figure~\ref{fig:pervasivesystem}. The methodology outcome is particularly interesting showing that the Early-Exit layer ($j=2$), being particularly demanding in terms of memory, is assigned to the Raspberry Pi 3B+ $n_{04}$ unit. When enough confidence is achieved at Early-Exit ($j=2$), the decision is directly sent from $n_{04}$ to the sink $f$ through $n_{06}$. Otherwise, the processing is forwarded from $n_{04}$ to $n_{01}$ to complete the processing up to $n_{08}$.
\subsection{The configuration with multiple CNNs}
\label{subsct:multipleCNNs}
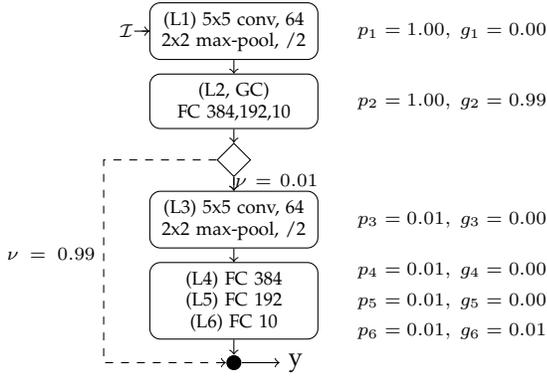
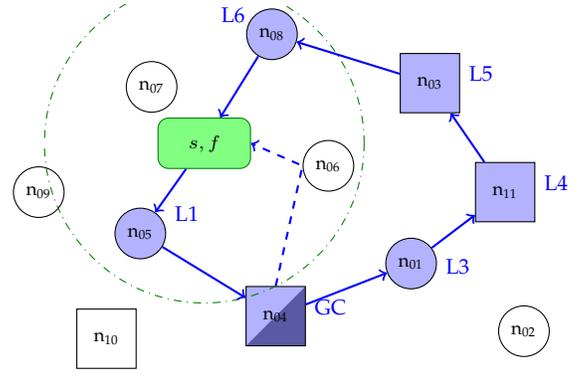
\begin{figure*}[!t]
	\centering
	\subfloat[][The 6-layer EX-CNN archicture with its $p_i$s and $g_i$s, for each $i \in \mathbb{N}_N$. $\mathcal{I}$ is the input image.]
	{
       	\begin{tikzpicture} [node distance = 0.05 cm, auto,align=center]%
			\centering
			\node (input) [text centered] {\scriptsize$\mathcal{I}$};
			\node (xgate1) [block, right = 0.1cm of input] {(L1) 5x5 conv, 64 \\ 2x2 max-pool, /2};
			\node (xgate1bis) [text centered, right=0.4 cm of xgate1] {\scriptsize$p_1=1.00,\ g_1=0.00$};
			\node (xgate2) [block, below = 0.18cm of xgate1] {\scriptsize(L2, GC)\\ FC 384,192,10};
			\node (xgate2bis) [text centered, right=0.4 cm of xgate2] {\scriptsize$p_2=1.00,\ g_2=0.99$};
			\node (decision) [diamond, draw,below = 0.18cm of xgate2] {};
			\node (xgate3) [block, below = 0.18cm of decision] {(L3) 5x5 conv, 64 \\ 2x2 max-pool, /2 };
			\node (xgate3bis) [text centered, right=0.4 cm of xgate3] {\scriptsize$p_3=0.01,\ g_3=0.00$};
			\node (xgate4) [block, below = 0.18cm of xgate3] {(L4) FC 384 \\ (L5) FC 192 \\ (L6) FC 10};
			\node (xgate4bis) [text centered, right=0.4 cm of xgate4] 
			{\scriptsize$p_4=0.01,\ g_4=0.00$\\
			 \scriptsize$p_5=0.01,\ g_5=0.00$\\
			 \scriptsize$p_6=0.01,\ g_6=0.01$};
			\node (output) [circle, fill=black,minimum size=0.2cm, inner sep=0pt,below = 0.2cm of xgate4] {};
			\node (outputText) [text centered, right = 0.5 cm of output] {y} ;
			%
			%
			%
			%
			%
			%
			\draw [->] (input) -- (xgate1);
            \draw (xgate1.180) -- ++(-0.2cm,0);
			\draw[->] (xgate1) -- (xgate2);
			\draw [->] (xgate2) -- (decision);
			\draw [dashed, ->] (decision.180) -- ++(-1.5, 0) |- node[xshift=-0.7cm, yshift=1.25cm, text width=2cm, font=\scriptsize]
			{$\nu=0.99$}
			(output);	
			\draw [dashed,->] (decision) -- node[xshift=-0.3cm, yshift=0.05cm, text width=1.5cm, font=\scriptsize]
			{$\nu=0.01$}
			(xgate3);
			\draw [->] (decision) -- (xgate3);
			\draw [->] (xgate3) -- (xgate4);
			\draw [->] (xgate4) -- (output);
			\draw [->] (output) -- (outputText);
		\end{tikzpicture}%
		\label{fig:6layergateclassificationcnn}
	}
	\hfill
	\subfloat[][The methodology outcome on the 6-layer EX-CNN on the IoT system shown in Figure~\ref{fig:pervasivesystem}.]
	{
        \begin{tikzpicture} [node distance = 0.05 cm]%
            \centering
            \clip (-2.9,-3.1) rectangle (5.2, 1.85);
            %

            \node at (0,0) [block, text width=3em,draw=green!50!black,fill=green!50!white] (source) {$s,f$};
            \node at (2.75,-1.6) [circle,draw,fill=blue!30!white] (n1) {\scriptsize n\textsubscript{01}};
            %
            \node at (4.25,-2.5) [circle,draw] (n2) {\scriptsize n\textsubscript{02}};
            \node at (3.,0.8) [square,draw,fill=blue!30!white] (n3) {\scriptsize n\textsubscript{03}};
            \node at (0.95,-2.3) [diagonal square={blue!30!gray}{blue!30!white},draw] (n4) {\scriptsize n\textsubscript{04}};
            \node at (-0.85,-1.2) [circle,draw,fill=blue!30!white] (n5) {\scriptsize n\textsubscript{05}};
            \node at (1.65,-0.3) [circle,draw] (n6) {\scriptsize n\textsubscript{06}};
            \node at (-0.7,0.75) [circle,draw] (n7) {\scriptsize n\textsubscript{07}};
            \node at (0.9,1.45) [circle,draw,fill=blue!30!white] (n8) {\scriptsize n\textsubscript{08}};
            \node at (-2.2,-0.65) [circle,draw] (n9) {\scriptsize n\textsubscript{09}};
            \node at (-1.3,-2.6) [square,draw] (n10) {\scriptsize n\textsubscript{10}};
            \node at (4,-0.65) [square,draw,fill=blue!30!white] (n11) {\scriptsize n\textsubscript{11}};

			\node [text centered, above right= -0.2 and 0.1cm of n5] {\footnotesize \textcolor{blue}{L1}};
			\node [text centered, above right= -0.5 and 0.0cm of n4] {\footnotesize\textcolor{blue}{GC}};
            \node [text centered, above right= -0.5 and 0.1cm of n1] {\footnotesize\textcolor{blue}{L3}};
			\node [text centered, above right= -0.5 and 0.0cm of n11] {\footnotesize\textcolor{blue}{L4}};
			\node [text centered, above right= -0.5 and 0.0cm of n3] {\footnotesize \textcolor{blue}{L5}};
			\node [text centered, above left=-0.2 and 0.0cm of n8] {\footnotesize \textcolor{blue}{L6}};
            %
            \draw (2.125,0) [dashdotted, color=green!50!black, radius=2.125cm, start angle=0, end angle=180] arc;
            \draw (2.125,0) [dashdotted, color=green!50!black, radius=2.125cm, start angle=0, end angle=-180] arc;
			\draw [thick,blue,->] (source) -- (n5);
			%
			\draw [thick,blue,->] (n5) -- (n4);
			\draw [thick,dashed,blue,->] (n4.90) -- (n6.180) -- (source.0);
			\draw [thick,blue,->] (n4) -- (n1);
			\draw [thick,blue,->] (n1) -- (n11);
			\draw [thick,blue,->] (n11) -- (n3);
			\draw [thick,blue,->] (n3) -- (n8);
			\draw [thick,blue,->] (n8) -- (source);
		\end{tikzpicture}
		\label{fig:singlegateclassificationcnnplacement}
	}
    \vspace*{-0.15cm}
	\caption{The methodology applied to a 6-layer EX-CNN, with $L=1$. Note that $d_{n_{04}, f}=2$, thus n\textsubscript{04} requires an intermediate hop, i.e., the node n\textsubscript{06}, to send the final classification.}
	\label{fig:singlegateclassificationcnnexample}
	%
\end{figure*}
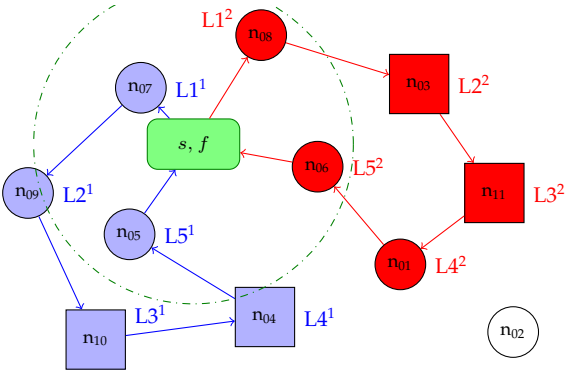
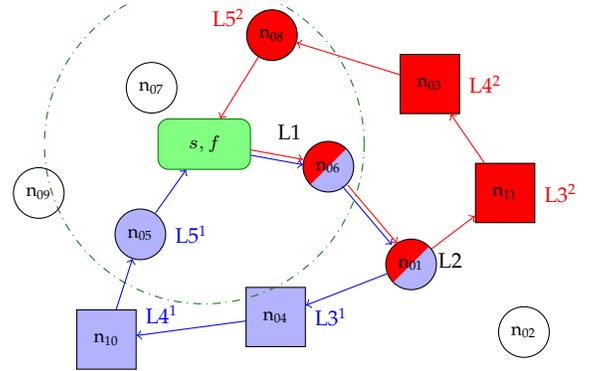
\begin{figure*}[!t]
	\centering
	%
	%
    \vspace*{-0.4cm}
	\subfloat[][The methodology outcome, with no shared layer between the two 5-layer CNNs (shown in Figure~\ref{fig:5layercnn}).]
	{
		\begin{tikzpicture} [node distance = 0.05 cm]%
        \centering
        \clip (-2.9,-3.1) rectangle (5.2, 1.85);
        %

        \node at (0,0) [block, text width=3em,draw=green!50!black,fill=green!50!white] (source) {$s,f$};
        \node at (2.75,-1.6) [circle,draw,fill=red] (n1) {\scriptsize n\textsubscript{01}};
        %
        \node at (4.25,-2.5) [circle,draw] (n2) {\scriptsize n\textsubscript{02}};
        \node at (3.,0.8) [square,draw,fill=red] (n3) {\scriptsize n\textsubscript{03}};
        \node at (0.95,-2.3) [square,draw,fill=blue!30!white] (n4) {\scriptsize n\textsubscript{04}};
        \node at (-0.85,-1.2) [circle,draw,fill=blue!30!white] (n5) {\scriptsize n\textsubscript{05}};
        \node at (1.65,-0.3) [circle,draw,fill=red] (n6) {\scriptsize n\textsubscript{06}};
        \node at (-0.7,0.75) [circle,draw,fill=blue!30!white] (n7) {\scriptsize n\textsubscript{07}};
        \node at (0.9,1.45) [circle,draw,fill=red] (n8) {\scriptsize n\textsubscript{08}};
        \node at (-2.2,-0.65) [circle,draw,fill=blue!30!white] (n9) {\scriptsize n\textsubscript{09}};
        \node at (-1.3,-2.6) [square,draw,fill=blue!30!white] (n10) {\scriptsize n\textsubscript{10}};
        \node at (4,-0.65) [square,draw,fill=red] (n11) {\scriptsize n\textsubscript{11}};
        \node [text centered, right= 0.0cm of n7] {\footnotesize\textcolor{blue}{L1\textsuperscript{1}}};
        \node [text centered, right= 0.0cm of n9] {\footnotesize\textcolor{blue}{L2\textsuperscript{1}}};
        \node [text centered, above right= -0.3 and 0.0cm of n10] {\footnotesize\textcolor{blue}{L3\textsuperscript{1}}};
        \node [text centered, right= 0.0cm of n4] {\footnotesize\textcolor{blue}{L4\textsuperscript{1}}};
        \node [text centered, right= 0.0cm of n5] {\footnotesize\textcolor{blue}{L5\textsuperscript{1}}};
        \node [text centered, above left= -0.25 and 0.0cm of n8] {\footnotesize\textcolor{red}{L1\textsuperscript{2}}};
        \node [text centered, right= 0.0cm of n3] {\footnotesize\textcolor{red}{L2\textsuperscript{2}}};
        \node [text centered, right= 0.0cm of n11] {\footnotesize\textcolor{red}{L3\textsuperscript{2}}};
        \node [text centered, right= 0.0cm of n1] {\footnotesize\textcolor{red}{L4\textsuperscript{2}}};
        \node [text centered, right= 0.0cm of n6] {\footnotesize\textcolor{red}{L5\textsuperscript{2}}};
        %
        \draw (2.125,0) [dashdotted, color=green!50!black, radius=2.125cm, start angle=0, end angle=180] arc;
        \draw (2.125,0) [dashdotted, color=green!50!black, radius=2.125cm, start angle=0, end angle=-180] arc;
		\draw [blue, ->] (source) -- (n7);
		\draw [blue,->] (n7) -- (n9);
		\draw [blue,->] (n9) -- (n10);
		\draw [blue,->] (n10) -- (n4);
		\draw [blue,->] (n4) -- (n5);
		\draw [blue,->] (n5) -- (source);
		\draw [red,->] (source) -- (n8);
		\draw [red,->] (n8) -- (n3);
		\draw [red,->] (n3) -- (n11);
		\draw [red,->] (n11) -- (n1);
		\draw [red,->] (n1) -- (n6);
		\draw [red,->] (n6) -- (source);
		\label{fig:multicnnnosharedlayers}
		\end{tikzpicture}
	}
	\hfill
	\subfloat[][The methodology outcome, with the first two layer shared between the two 5-layer CNNs.]
	{
		\begin{tikzpicture} [node distance = 0.05 cm, auto,align=center]%
            \centering
            \clip (-2.9,-3.1) rectangle (5.2, 1.85);
            %

            \node at (0,0) [block, text width=3em,draw=green!50!black,fill=green!50!white] (source) {$s,f$};
            \node at (2.75,-1.6) [diagonal circle={blue!30!white}{red},draw] (n1) {\scriptsize n\textsubscript{01}};
            %
            \node at (4.25,-2.5) [circle,draw] (n2) {\scriptsize n\textsubscript{02}};
            \node at (3.,0.8) [square,draw,fill=red] (n3) {\scriptsize n\textsubscript{03}};
            \node at (0.95,-2.3) [square,draw,fill=blue!30!white] (n4) {\scriptsize n\textsubscript{04}};
            \node at (-0.85,-1.2) [circle,draw,fill=blue!30!white] (n5) {\scriptsize n\textsubscript{05}};
            \node at (1.65,-0.3) [diagonal circle={blue!30!white}{red},draw] (n6) {\scriptsize n\textsubscript{06}};
            \node at (-0.7,0.75) [circle,draw] (n7) {\scriptsize n\textsubscript{07}};
            \node at (0.9,1.45) [circle,draw,fill=red] (n8) {\scriptsize n\textsubscript{08}};
            \node at (-2.2,-0.65) [circle,draw] (n9) {\scriptsize n\textsubscript{09}};
            \node at (-1.3,-2.6) [square,draw,fill=blue!30!white] (n10) {\scriptsize n\textsubscript{10}};
            \node at (4,-0.65) [square,draw,fill=red] (n11) {\scriptsize n\textsubscript{11}};
            \node [text centered, above left = -0.0cm and 0.0cm of n6] {\footnotesize L1};
            \node [text centered, above right= -0.4cm and 0.0cm of n1] {\footnotesize L2};
            \node [text centered, right= 0.0cm of n4] {\footnotesize\textcolor{blue}{L3\textsuperscript{1}}};
            \node [text centered, above right= -0.3 and 0.0cm of n10] {\footnotesize\textcolor{blue}{L4\textsuperscript{1}}};
            \node [text centered, right= 0.0cm of n5] {\footnotesize\textcolor{blue}{L5\textsuperscript{1}}};
            \node [text centered, right= 0.0cm of n11] {\footnotesize\textcolor{red}{L3\textsuperscript{2}}};
            \node [text centered, right= 0.0cm of n3] {\footnotesize\textcolor{red}{L4\textsuperscript{2}}};
            \node [text centered, above left= -0.25 and 0.0cm of n8] {\footnotesize\textcolor{red}{L5\textsuperscript{2}}};
            %
            \draw (2.125,0) [dashdotted, color=green!50!black, radius=2.125cm, start angle=0, end angle=180] arc;
            \draw (2.125,0) [dashdotted, color=green!50!black, radius=2.125cm, start angle=0, end angle=-180] arc;
			\draw [blue, ->,transform canvas={yshift=-1pt}] (source) -- (n6);
			\draw [blue,->,transform canvas={xshift=-1pt}] (n6) -- (n1);
			\draw [blue,->] (n1) -- (n4);
			\draw [blue,->] (n4) -- (n10);
			\draw [blue,->] (n10) -- (n5);
			\draw [blue,->] (n5) -- (source);
			\draw [red,->,transform canvas={yshift=+1pt}] (source) -- (n6);
			\draw [red,->,transform canvas={xshift=+1pt,yshift=+0.5pt}] (n6) -- (n1);
			\draw [red,->] (n1) -- (n11);
			\draw [red,->] (n11) -- (n3);
			\draw [red,->] (n3) -- (n8);
			\draw [red,->] (n8) -- (source);
	 		\label{fig:multicnnsharedlayers}
		\end{tikzpicture}
	}
	\caption{The methodology applied to two 5-layer CNNs  and the IoT system in Figure~\ref{fig:pervasivesystem} with $L=1$.}
	\label{fig:multicnn}
	%
\end{figure*}
An interesting application scenario is the one with multiple CNNs, either without or with shared processing layers. EX-CNNs have not been considered here, hence $p_{u,j}=1$, for each $u \in \mathbb{N}_C$ and $j \in \{1,\ldots,M_u\}$.

In this configuration, the objective function modelling the latency in decision making becomes:
\begin{equation}
    \sum_{u=1}^{C}\sum_{i=1}^{N}\sum_{k=i}^{N}\sum_{j=1}^{M-1} \alpha_{u,i,j}\cdot\alpha_{u,k,j+1}\cdot \frac{K_{u,j}}{\rho} \cdot d_{i,k}
    + \sum_{i=1}^{N} t_{i}^{(p)} + t_{s} + t_{f},
    \label{eq:objectivefunctiondistancemulticnns}
\end{equation}
with constraints defined in Eqs.~\eqref{eq:nodeconstraint},~\eqref{eq:memoryconstraint},~\eqref{eq:computationconstraint} and~\eqref{eq:assignedlayerconstraint}, and where the source time $t_s$, the processing time $t_i^{(p)}$, and the sink time $t_f$ are modified as follow:
\begin{alignat}{2}
    &t_{s}&&=\sum_{u=1}^{C} \sum_{i=1}^{N} \alpha_{u, i,1} \cdot \frac{K_{u,s}}{\rho} \cdot d_{s,i},
    \label{eq:multicnnsourcetime} \\
    &t_{i}^{(p)}&&= \sum_{u=1}^{C}\sum_{j=1}^{M} \alpha_{u,i,j} \cdot \frac{c_{u, j}}{e_i},
    \label{eq:multicnnprocessingtime} \\
    &t_{f}&&= \sum_{u=1}^{C}\sum_{i=1}^{N} \alpha_{u, i,M_u} \cdot \frac{K_{u,M_u}}{\rho} \cdot d_{i,f}.
    \label{eq:multicnnsinktime}
\end{alignat}

Finally, to deal with shared layers, the constraint defined in  Eq.~\eqref{eq:sharedlayerequalplacemet} is introduced per each shared layer, whereas the constraints in  Eqs.~\eqref{eq:nodeconstraint},~\eqref{eq:memoryconstraint}, and~\eqref{eq:computationconstraint} are modified accordingly, as detailed in Section~\ref{sct:theMethodology} with Eqs.~\eqref{eq:sharedlayrenodeconstraint},~\eqref{eq:sharedlayermemoryconstraint}, and~\eqref{eq:sharedlayercomputationconstraint}.
As an example, it is provided the complete extension to deal with the two first shared layers for the case shown in Figure~\ref{fig:multicnnsharedlayers}. The additional constraints are defined in Eqs.~\eqref{eq:commonlayerauxiliaryvariablesequal_example1} and~\eqref{eq:commonlayerauxiliaryvariablesequal_example2} to ensure that the shared layers $j=1$ and $j=2$ (of CNNs $u=1$ and $u=2$) are assigned to the same node $i$. Then, the Eqs.~\eqref{eq:nodeconstraint},~\eqref{eq:memoryconstraint}, and~\eqref{eq:computationconstraint} are modified as follow:
\begin{align}
    &\forall i \in \mathbb{N}_N&&\sum_{u=1}^{C}\sum_{j=1}^{M} \alpha_{u,i,j}
    \le 
    L + \alpha_{1,i,1} + \alpha_{1,i,2},
    \label{eq:sharedlayrenodeconstraint_supplementary} \\
    &\forall i \in \mathbb{N}_N&&\sum_{u=1}^{C}\sum_{j=1}^{M} \alpha_{u,i,j} \cdot m_{u,j} 
    \le 
    \bar{m}_i + \hat{m},
    \label{eq:sharedlayermemoryconstraint_supplementary} \\
    &\forall i \in \mathbb{N}_N&&\sum_{u=1}^{C}\sum_{j=1}^{M} \alpha_{u,i,j} \cdot c_{u,j} 
    \le 
    \bar{c}_i + \hat{c}.
    \label{eq:sharedlayercomputationconstraint_supplementary}
\end{align}
where
\begin{align}
    \hat{m} &=  \alpha_{1,i,1}\cdot m_{1,1} + \alpha_{1,i,2}\cdot m_{1,2} 
    \label{eq:sharedlayermemoryconstraint_mhat}\\
    \hat{c} &= \alpha_{1,i,1}\cdot c_{1,1} + \alpha_{1,i,2}\cdot c_{1,2}
    \label{eq:sharedlayercomputationalconstraint_chat}
\end{align}

It is noteworthy to point out that there is no difference in defining Eqs.~\eqref{eq:sharedlayrenodeconstraint_supplementary},~\eqref{eq:sharedlayermemoryconstraint_supplementary}, and~\eqref{eq:sharedlayercomputationconstraint_supplementary} with the variables of the CNN $u=2$, instead of those of CNN $u=1$, as done here. It is indeed sufficient to relax the constraints with $k-1$ variables out of $k$, where $k$ represents the number of CNNs a layer is shared among, in order to count that shared layer only once.

The proposed methodology has been applied to two instances of the 5-layer CNN described in Figure~\ref{fig:5layercnn} without common processing layers, operating in the IoT system depicted in Figure~\ref{fig:pervasivesystem} and with $L=1$.
Interestingly, the outcome of the methodology, depicted in Figure~\ref{fig:multicnnnosharedlayers}, shows that the placement of both CNNs represent the optimal solution of the single-CNN configuration.
Moreover, the methodology has been applied to the case where the convolutional layers $L1$ and $L2$ are shared between the two CNNs. This solution is inspired by the transfer learning paradigm where two CNNs might share low-level representation processing layers, while high-level ones are specific for each CNN. As described in Section~\ref{sct:theMethodology}, the following constraints need to be added to the optimization problem:
\begin{align}
    \label{eq:commonlayerauxiliaryvariablesequal_example1}
    \forall i \in \mathbb{N}_N\qquad &\alpha_{1,i,1} = \alpha_{2,i,1},  \\
    \forall i \in \mathbb{N}_N\qquad &\alpha_{1,i,2} = \alpha_{2,i,2},
    \label{eq:commonlayerauxiliaryvariablesequal_example2}
\end{align}
and the constraints Eq.~\eqref{eq:sharedlayrenodeconstraint},~\eqref{eq:sharedlayermemoryconstraint} and~\eqref{eq:sharedlayercomputationconstraint} have to be redefined accordingly. The methodology outcome in this scenario is interesting showing that common layers $L1$ and $L2$ have been placed in IoT units $n_{06}$ and $n_{01}$, respectively, while, after $n_{01}$, the processing takes two different paths.
\subsection{Open Points}
\label{subsct:complexityopenpoints}
Currently, the proposed methodology neither takes into account the energy status of IoT units nor network failures~\cite{farhan2017survey}.  The energy status can be managed by introducing a constraint for each IoT unit depending on the remaining energy value, e.g., forcing the variables $\alpha_{\cdot,i,\cdot}$s of a low-energy node $i$ to zero or introducing penalties for assigning a layer to those nodes. The network failures could be managed by modifying the transmission time defined in Eq.~\eqref{eq:transmissiontime} as $(1 + \xi^{i,j}) \cdot K_{u,j} / \rho \cdot d_{i,k} $ taking into account a probability of retransmission $\xi^{i,j}$ for each pair of nodes $(i,j)$.

It is noteworthy to point out that, thanks to the transfer learning paradigm, the hierarchy of layers of the CNNs can be considered as general feature extractors~\cite{yosinski2014transferable}. For this reason, the deployed CNNs can be easily reconfigured to address a different image-classification problem by replacing only upper layers.  
Moreover, this optimization phase can be scheduled periodically or when needed to manage variations in the IoT network configuration (e.g., due to the removal or insertion of IoT units). This is a crucial aspect in the scenario of mobile IoT units, a case that is not considered in this paper. In fact, thanks to the  transfer learning approach and by periodically recomputing the CNN allocation, the methodology could be applied to units changing their position in the environments they are operating in.
%
\section{Experimental Results}
\label{sct:experimentalResults}
The proposed methodology has been validated on five CNNs and four families of off-the-shelf IoT devices in a synthetic scenario of distributed image classification for the control of a critical area (e.g., recognition of the presence of target objects in a given area through image classification). 
The monitored area is assumed to be a 30m square and the positions of the IoT units as well as those of the sources $s_u$s and the sink $f$ are randomly selected following a uniform distribution. 
The hyper-parameter $L$, setting the maximum number of CNN layers per IoT unit, ranges from 1 to 5. 

The rest of the section is organized as follows. Section~\ref{subsct:cnn} details the considered CNNs, Section~\ref{subsct:nodes} describes the families of considered off-the-shelf IoT units and their transmission technologies, and Section~\ref{subsct:figuresofmerit} describes the figure of merit. Sections~\ref{subsct:firstiotsystem} and~\ref{subsct:harsherIoTsystem} describe the experimental results in two different IoT systems.
\subsection{The considered CNNs}
\label{subsct:cnn}
The first two CNNs are the 5-layer CNN and the 6-layer EX-CNN shown in Figures~\ref{fig:5layercnn} and~\ref{fig:6layergateclassificationcnn}, respectively. These two CNNs, whose values of $m_j$s, $c_j$s, and $K_j$s are detailed in Table~\ref{tab:6layergateclassification}, receive in input a 28x28 RGB image and have the following processing layers: two convolutional (with 64 5x5 filters) followed by a 2x2 maximum pooling with stride 2, and three ful\-ly-con\-nec\-ted layers with 384, 192 and 10 neurons, respectively. In the 6-layer EX-CNN, the Early-Exit is after the first pooling layer and is composed by three ful\-ly-con\-nec\-ted layers with 384, 192 and 10 neurons.

The third CNN is the AlexNet~\cite{krizhevsky2012imagenet}, whose details are given in Table~\ref{tab:alexnet}. Such CNN works on 227x227 RGB images and is endowed with 5 convolutional layers (with 96 11x11, 256 5x5, 384 3x3, 384 3x3 and 256 3x3 filters, respectively) and 3 ful\-ly-con\-nec\-ted layers with 4096, 4096 and 2 neurons. In addition, 3x3 maximum pooling layers with stride 2 are present after the first, second and fifth convolutional layers. An Early-Exit variant of the AlexNet~\cite{disabato2018reducing} has been considered where the Early-Exit is placed after the second maximum pooling layer and is composed of three ful\-ly-con\-nec\-ted layers with 128, 64 and 2 neurons, respectively.

The fifth considered CNN is the ResNet 101~\cite{he2016deep}, whose details are given in Table~\ref{tab:resnet}. This CNN works on 224x224 RGB images and is composed by a sequence of Bottleneck blocks (i.e., a sequence of a 1x1 convolution, a 3x3 convolution, and a 1x1 convolution with 4 times filters w.r.t to the other two convolutions) of increasing number of filters. The last two layers are an average pooling and a ful\-ly-con\-nec\-ted.

In all the considered CNNs, the ReLUs, the batch normalization, and the softmax layers have not been explicitly mentioned since they have no parameter to store and negligible computational demands.
\subsection{The considered IoT Units}
\label{subsct:nodes}
\begin{table}[!t]
	\centering
	\scriptsize
	\caption{The memory demand $m_j$, the computational load $c_j$, and the memory $K_j$ required to store the intermediate representations of the ResNet 101~\cite{he2016deep}, with a 4B data type.
    When there are two values for $m_j$ and $c_j$ in repeated sequence of layers, the former one refers to the first repetition, the latter to all the subsequent repetitions.}
 \resizebox{\linewidth}{!}{
	\begin{tabular}{C{0.5cm}@{\hskip2pt}@{\hskip2pt}m{2.5cm}@{\hskip2pt}@{\hskip2pt}C{2cm}@{\hskip2pt}@{\hskip2pt}C{2cm}@{\hskip2pt}@{\hskip2pt}C{2cm}@{}m{0pt}@{}}
		&Layer ($j$)&$m_j$ (KB)&$c_j$ ($10^6$ mult.)&$K_j$ (KB)&\\ \midrule[1.25pt]
		s&Source  (Im. 224x224x3)&-&-&602.12& \\ \midrule
		$1_1$&7x7 conv, 64, /2&37.63 &118.01&3\,211.26& \\
		$1_2$&3x3 max pool, /2&-&1.81&802.82& \\ \midrule
        
        &\multirow{3}{2.44cm}{3x\resizebox*{!}{0.875cm}{$\begin{cases}
            \text{1x1 conv, 64}\\
            \text{3x3 conv, 64}\\
            \text{1x1 conv, 256}
            \end{cases}$}}&16.38--65.54&12.85--51.38&802.82&\\
		2&&147.46&115.61&802.82& \\ 
        &&65.54&51.38&3\,211.26& \\ \cmidrule{3-6}
        &&786.43&616.56&3\,211.26&\\\midrule
        
        &\multirow{3}{2.44cm}{4x\resizebox*{!}{0.875cm}{$\begin{cases}
                \text{1x1 conv, 128}\\
                \text{3x3 conv, 128}\\
                \text{1x1 conv, 512}
                \end{cases}$}}&131.07--262.14&25.69--51.38&401.41&\\
        3&&65.54&115.61&401.41& \\ 
        &&262.14&51.38&1\,605.63& \\\cmidrule{3-6}
        &&2228.2&757.86&1\,605.63&\\\midrule

        &\multirow{3}{2.44cm}{5x\resizebox*{!}{0.875cm}{$\begin{cases}
                \text{1x1 conv, 256}\\
                \text{3x3 conv, 256}\\
                \text{1x1 conv, 1024}
                \end{cases}$}}&524.29--1\,048.58&25.69--51.38&200.71&\\
        4&&2\,359.30&115.61&200.71& \\ 
        &&1\,048.58&51.38&800.82& \\ \cmidrule{3-6}
        &&21\,757.95&950.53&800.82&\\\midrule
        
        &\multirow{3}{2.44cm}{6x\resizebox*{!}{0.875cm}{$\begin{cases}
                \text{1x1 conv, 256}\\
                \text{3x3 conv, 256}\\
                \text{1x1 conv, 1024}
                \end{cases}$}}&1\,048.58&51.38&200.71&\\
        5,6,7&&2\,359.30&115.61&200.71& \\ 
        &&1\,048.58&51.38&800.82& \\ \cmidrule{3-6}
        &&26\,738.69&1\,156.06&800.82&\\\midrule
        
%
        
        &\multirow{3}{2.44cm}{3x\resizebox*{!}{0.875cm}{$\begin{cases}
                \text{1x1 conv, 512}\\
                \text{3x3 conv, 512}\\
                \text{1x1 conv, 2048}
                \end{cases}$}}&2\,097.15--4\,194.30&25.69--51.38&100.35&\\
        8&&9\,437.18&115.61&200.71& \\ 
        &&4\,194.30&51.38&800.82& \\ \cmidrule{3-6}
        &&51\,380.22&629.41&800.82&\\\midrule
        
        $9_1$&7x7 avg pool&-&0.10&8.19& \\ 
        $9_2$&fc 1000&8\,388.61&2.08&4& \\ \bottomrule[1.25pt]
	\end{tabular}
}
	\label{tab:resnet}
\end{table}
%
%
%
%
\begin{table}[!t]
	\centering
	\scriptsize
	\caption{The memory demand $m_j$, the computational load $c_j$, and the memory $K_j$ required to store the intermediate representations of four (EX--)CNNs, with a 4B data type and the Early-Exit layer marked with an asterisk. In that layer, $K_j$ represents the dimensions of the representation sent to the layer $j+1$ when the classification is not taken at layer $j$.}
	%
 \subfloat[][The 5-layer CNN and the 6-layer EX-CNN.]{
	\begin{tabular}{C{0.5cm}@{\hskip2pt}@{\hskip2pt}m{2.5cm}@{\hskip2pt}@{\hskip2pt}C{1.85cm}@{\hskip2pt}@{\hskip2pt}C{1.6cm}@{\hskip2pt}@{\hskip2pt}C{1.6cm}@{}m{0pt}@{}}
		&Layer ($j$)&$m_j$ (KB)&$c_j$ ($10^6$ mult.)&$K_j$ (KB)&\\ \midrule[1.25pt]
		s&Source (Im. 28x28x3)&-&-&9.41& \\ \midrule 
		$1_1$&5x5 conv, 64&19.20&3.76&200.70& \\
		$1_2$&2x2 pool, /2&-&0.05&50.18& \\ \midrule
		2*&gc1 (fc 384,192,10)&19\,570.18&4.89&50.18& \\ \midrule
		$3_1$&5x5 conv, 64&409.60&20.07&50.18& \\
		$3_2$&2x2 pool, /2&-&0.01&12.54& \\ \midrule
		4&fc 384&4\,816.90&1.20&1.54& \\ 
		5&fc 192&294.91&0.07&0.77& \\ 
		6&fc 10&7.68&$2\cdot10^{-3}$&0.04& \\ \bottomrule[1.25pt]
	\end{tabular}
	%
	\label{tab:6layergateclassification}
}

\subfloat[][The AlexNet~\cite{krizhevsky2012imagenet} and its Early-Exit version~\cite{disabato2018reducing}.] {
    \begin{tabular}{C{0.5cm}@{\hskip2pt}@{\hskip2pt}m{2.5cm}@{\hskip2pt}@{\hskip2pt}C{1.85cm}@{\hskip2pt}@{\hskip2pt}C{1.6cm}@{\hskip2pt}@{\hskip2pt}C{1.6cm}@{}m{0pt}@{}}
        &Layer ($j$)&$m_j$ (KB)&$c_j$ ($10^6$ mult.)&$K_j$ (KB)&\\ \midrule[1.25pt]
        s&Source (Im. 227x227x3)&-&-&618.35& \\ \midrule
        $1_1$&11x11 conv, 96, /4&139.78&105.42&1161.60& \\
        $1_2$&3x3 pool, /2&-&0.31&279.94& \\ \midrule
        $2_1$&5x5 conv, 256&1\,229.82&223.95&746.50& \\
        $2_2$&3x3 pool, /2&-&0.39&173.06& \\ \midrule
        3\textsuperscript{*}&gc1(fc 128,64,2)&22185.22&5.55&173.06& \\ \midrule
        4&3x3 conv, 384&3\,540.48&149.52&259.58& \\ 
        5&3x3 conv, 384&2\,655.74&112.14&259.58& \\ 
        $6_1$&3x3 conv, 256&1\,770.50&74.76&173.06& \\
        $6_2$&3x3 pool, /2&-&0.08 &36.86& \\ \midrule
        7&fc 4096&151\,011.39&37.75&16.38& \\ 
        8&fc 4096, 2&67\,158.02&16.78&0.01& \\ \bottomrule[1.25pt]
    \end{tabular}
    %
    \label{tab:alexnet}
}
\end{table}
In this experimental section we considered four  families of IoT units, whose technological details are given in Table~\ref{tab:nodes}: the STM32H7, a simple IoT unit endowed with a 400 MHz-Cortex M7 and 1 MB of RAM; the Raspberry Pi 3B+, a more powerful IoT units endowed with a 1.4GHz 64-bit quad-core processor and 1GB of RAM; the OrangePi Zero, endowed with a 800MHz to 1.2GHz quad-core Cortex-A7 and 256 MB of RAM;  and the BeagleBone AI having a 1.5GHz dual-core ARM Cortex--A15 and 1GB of RAM.

The maximum memory usage $\bar{m}_i$s has been defined as half of the available RAM memory, i.e., $512$KB for the STM32H7, $128$MB for the OrangePi Zero, and $512$MB for the both the BeagleBone AI and the Raspberry Pi3B+. The number of multiplications per second $e_i$s has been defined as a tenth of the clock cycles (per number of cores). Hence, $e=40$ for the STM32H7, $e=300$ for the BeagleBone AI ($150$ per core), $e=480$ for the OrangePi Zero ($120$ per core, if we consider the maximum frequency of 1.2GHz), and $e=560$ for the Raspberry 3B+ ($140$ per core). 
The constraints on the computational load $\bar{c}_i$s have not been considered since they are application-specific.

The transmission technologies the IoT units are equipped with are the \textit{Wi-Fi 4} (standard IEEE 802.11n) and \textit{Wi-Fi HaLow} (standard IEEE 802.11n). The transmission range is $d_t = 7.5m$ (a tenth of the minimum indoor range). The Wi-Fi 4 data-rate is $\rho=72.2$ Mb/s, that corresponds to the single-antenna scenario with 64-QAM modulation on the 20 MHz channel~\cite{xiao2005ieee}, whereas the \textit{Wi-Fi HaLow} one is $\rho=7.2$ Mb/s with a single-antenna and 64-QAM modulation on the 2 MHz channel~\cite{adame2014ieee}.
\subsection{Figures of Merit}
\label{subsct:figuresofmerit}
The proposed methodology is evaluated on the ``data production to decision making''-latency $t$ defined as the time between image acquisition and classification reception at $f$. To further clarify the effects of data communication and computation, $t$ is split into the transmission $t_t$ and the processing $t_p$ terms. The former term refers to the sum of all transmission times (from a source to IoT units, between IoT units, or from IoT units to the target unit $f$); the latter sums the processing times on the IoT units.
These terms are computed as defined in Section~\ref{sct:theMethodology}, whereas additional sources of delays, such as transmission handshakes or repeated transmissions (due to failures) have been neglected.
For each setting, transmission technology, and configuration, the evaluated figure of metric is the mean $\pm$ standard deviation of each latency term, i.e., $t$, $t_t$, and $t_p$, computed on $500$ randomly generated IoT systems.
It is noteworthy to point out that the accuracy has not been considered as a metric since the proposed method does not introduce any approximation w.r.t the original CNN, hence there is no accuracy loss due to placement of the CNN layers.
\subsection{The First IoT System: 30 IoT Units and Two Technological Families}
\label{subsct:firstiotsystem}
The first  IoT system comprises $N=30$ IoT units belonging to two technological families, i.e., the STM32H7 and the Raspberry Pi 3B+, with three settings for the IoT units partitioning, i.e., 10\%--90\%, 50\%--50\% and 90\%--10\%.
\subsubsection{Single-CNN Configuration}
\label{subsct:resultssinglecnn}
\begin{table}[!t]
	\centering
	\scriptsize
	\caption{The maximum memory usage $\bar{m}_i$ (defined as half of the available RAM), 
		and the million ($10^6$) multiplications per second $e_i$s (defined as a tenth of the clock cycles performed in one second) of a few off-the-shelf IoT units.}
	\begin{tabular}{C{0.5cm}m{2.7cm}@{\hskip3pt}@{\hskip3pt}C{2cm}@{\hskip3pt}@{\hskip3pt}C{2cm}@{}m{0pt}@{}}
		&Node ($i$)&$\bar{m}_i$ (KB)&$e_i$ ($10^6$ mult.)&\\ \midrule[1.25pt]
		S1&STM32H7&512&40& \\ 
        B1&BeagleBone AI&524\,288&360&\\
        O1&OrangePi Zero&131\,072&480&\\
		R1&Raspberry Pi 3B+&524\,288&560& \\ 
		\bottomrule[1.25pt]
	\end{tabular}
	\label{tab:nodes}
\end{table}

\begin{table*}[t]
    \centering
    \scriptsize
    \caption{Single-(EX--)CNN configuration results with $N=30$ STM32H7 and Raspberry Pi 3B+ units in the 50\%-50\% scenario. The figure of merit (mean $\pm$ std) is the latency $t$, i.e., the transmission time $t_t$ plus the processing time $t_p$.}
    %
   \subfloat[][The results with the Wi-Fi 4 as adopted transmission technology.]{
    \begin{tabular}{C{0.4cm}@{\hskip2pt}@{\hskip2pt}C{0.40cm}@{\hskip2pt}@{\hskip2pt}C{2.1cm}@{\hskip2pt}@{\hskip2pt}C{2.1cm}@{\hskip2pt}@{\hskip2pt}C{2.1cm}cC{0.4cm}@{\hskip2pt}@{\hskip2pt}C{0.40cm}@{\hskip2pt}@{\hskip2pt}C{2.1cm}@{\hskip2pt}@{\hskip2pt}C{2.1cm}@{\hskip2pt}@{\hskip2pt}C{2.1cm}@{}m{0pt}}
        &&\multicolumn{3}{c}{Latency $t$ (ms)}&&&&\multicolumn{3}{c}{Latency $t$ (ms)}&\\ \cmidrule{3-5}\cmidrule{9-11}
        &L&$t_t$&$t_p$&$t=t_t+t_p$&&&L&$t_t$&$t_p$&$t=t_t+t_p$&\\ \cmidrule[1.25pt]{1-5}\cmidrule[1.25pt]{7-12}
        \multirow{6}{0.4cm}{\rotatebox{90}{\centering 5-layer CNN}}
        &1&$7.49\pm0.36$&$44.93\pm0.00$&$52.42\pm0.36$&&\multirow{6}{0.4cm}{\rotatebox{90}{\centering AlexNet}}
        &1&$203.06\pm36.85$&$1257.71\pm0.00$&$1460.77\pm36.85$&\\  
        &2&$1.78\pm0.21$&$44.93\pm0.00$&$46.71\pm0.21$&&&2&$127.81\pm28.49$&$1257.71\pm0.00$&$1385.52\pm28.49$& \\  
        &3&$0.48\pm0.14$&$44.93\pm0.00$&$45.41\pm0.14$&&&3&$98.75\pm26.63$&$1257.71\pm0.00$&$1356.46\pm26.63$&\\  
        &4&$0.37\pm0.11$&$44.93\pm0.00$&$45.30\pm0.11$&&&4&$95.82\pm26.54$&$1257.71\pm0.00$&$1353.53\pm26.54$& \\  \cmidrule{2-5}\cmidrule{8-12}
        &C&$0.28\pm0.11$&$44.93\pm0.00$&$45.21\pm0.11$&&&C&$72.49\pm24.08$&$1257.71\pm0.00$&$1330.20\pm24.08$&\\ \cmidrule[1.125pt]{1-5}\cmidrule[1.125pt]{7-12}
        \multirow{6}{0.4cm}{\rotatebox{90}{\tiny\centering 6-layer EX-CNN}}
        &1&$5.87\pm0.27$&$7.27\pm0.00$&$13.14\pm0.27$&&\multirow{6}{0.4cm}{\rotatebox{90}{\centering EX-AlexNet}}
        &1&$129.42\pm28.29$&$598.92\pm81.81$&$728.35\pm86.17$&\\  
        &2&$0.34\pm0.10$&$7.27\pm0.00$&$7.61\pm0.10$&&&2&$93.70\pm24.49$&$596.19\pm0.00$&$689.89\pm24.49$& \\  
        &3&$0.29\pm0.10$&$7.27\pm0.00$&$7.57\pm0.10$&&&3&$72.10\pm22.08$&$596.19\pm0.00$&$668.29\pm22.08$& \\  
        &4&$0.28\pm0.10$&$7.27\pm0.00$&$7.55\pm0.10$&&&4&$72.07\pm22.08$&$596.19\pm0.00$&$668.26\pm22.08$&\\  \cmidrule{2-5}\cmidrule{8-12}
        &C&$0.28\pm0.10$&$7.27\pm0.00$&$7.55\pm0.10$&&&C&$71.84\pm22.07$&$596.19\pm0.00$&$668.03\pm22.07$&\\ \cmidrule[1.25pt]{1-5}\cmidrule[1.25pt]{7-12}
    \end{tabular}
}

\subfloat[][The results with the Wi-Fi HaLow as adopted transmission technology.]{
        \begin{tabular}{C{0.4cm}@{\hskip2pt}@{\hskip2pt}C{0.40cm}@{\hskip2pt}@{\hskip2pt}C{2.1cm}@{\hskip2pt}@{\hskip2pt}C{2.1cm}@{\hskip2pt}@{\hskip2pt}C{2.1cm}cC{0.4cm}@{\hskip2pt}@{\hskip2pt}C{0.40cm}@{\hskip2pt}@{\hskip2pt}C{2.1cm}@{\hskip2pt}@{\hskip2pt}C{2.1cm}@{\hskip2pt}@{\hskip2pt}C{2.1cm}@{}m{0pt}}
        &&\multicolumn{3}{c}{Latency $t$ (ms)}&&&&\multicolumn{3}{c}{Latency $t$ (ms)}&\\ \cmidrule{3-5}\cmidrule{9-11}
        &L&$t_t$&$t_p$&$t=t_t+t_p$&&&L&$t_t$&$t_p$&$t=t_t+t_p$&\\ \cmidrule[1.25pt]{1-5}\cmidrule[1.25pt]{7-12}
        \multirow{6}{0.4cm}{\rotatebox{90}{\centering 5-layer CNN}}
        &1&$74.82\pm3.94$&$45.16\pm0.57$&$119.98\pm4.06$&&\multirow{6}{0.4cm}{\rotatebox{90}{\centering AlexNet}}
        &1&$2042.48\pm401.64$&$1265.89\pm141.47$&$3308.37\pm432.91$&\\ 
        &2&$17.75\pm2.08$&$44.93\pm0.00$&$62.68\pm2.08$&&&2&$1298.23\pm332.66$&$1263.16\pm115.58$&$2561.39\pm361.81$& \\  
        &3&$4.49\pm0.89$&$45.08\pm0.46$&$49.57\pm1.03$&&&3&$1009.79\pm334.87$&$1260.44\pm81.77$&$2270.22\pm351.10$& \\  
        &4&$3.64\pm0.87$&$44.93\pm0.00$&$48.57\pm0.87$&&&4&$976.71\pm317.48$&$1263.16\pm115.58$&$2239.87\pm348.56$&\\  \cmidrule{2-5}\cmidrule{8-12}
        &C&$2.80\pm0.88$&$44.93\pm0.00$&$47.73\pm0.88$&&&C&$750.02\pm322.44$&$1260.44\pm81.77$&$2010.46\pm339.52$& \\ \cmidrule[1.25pt]{1-5}\cmidrule[1.25pt]{7-12}
        \multirow{6}{0.4cm}{\rotatebox{90}{\tiny\centering 6-layer EX-CNN}}
        &1&$58.72\pm2.47$&$7.27\pm0.01$&$65.99\pm2.47$&&\multirow{6}{0.4cm}{\rotatebox{90}{\centering EX-AlexNet}}
        &1&$1304.74\pm310.87$&$598.93\pm81.91$&$1903.66\pm324.56$&\\  
        &2&$3.53\pm1.15$&$7.27\pm0.00$&$10.80\pm1.15$&&&2&$947.04\pm258.12$&$596.19\pm0.00$&$1543.23\pm258.12$& \\  
        &3&$3.04\pm1.10$&$7.27\pm0.00$&$10.31\pm1.10$&&&3&$728.19\pm221.12$&$596.19\pm0.00$&$1324.38\pm221.12$&\\  
        &4&$2.90\pm1.10$&$7.27\pm0.00$&$10.17\pm1.10$&& &4&$727.87\pm221.10$&$596.19\pm0.00$&$1324.06\pm221.10$& \\  \cmidrule{2-5}\cmidrule{8-12}
        &C&$2.88\pm1.10$&$7.27\pm0.00$&$10.16\pm1.10$&&&C&$725.53\pm220.95$&$596.19\pm0.00$&$1321.72\pm220.95$&\\ \cmidrule[1.25pt]{1-5}\cmidrule[1.25pt]{7-12}
    \end{tabular}
}
    \label{tab:singlecnnresults}
\end{table*}
%
%
\begin{table*}[t]
	\centering
	\scriptsize
	\caption{The multi CNN configuration results with $N=30$ STM32H7 and Raspberry Pi 3B+ units and two 5-layer CNNs, in the three scenarios, with the Wi-Fi 4 and Wi-Fi Halow transmission technologies. The figure of merit (mean $\pm$ std) is the latency $t$, i.e., the transmission time $t_t$ plus the processing time $t_p$ and is summed over all the CNNs.}
\subfloat[][The results with the Wi-Fi 4 as adopted transmission technology.]{

		\begin{tabular}{@{}m{0pt}@{}C{1.1cm}@{\hskip2pt}@{\hskip2pt}C{0.4cm}@{\hskip2pt}@{\hskip2pt}C{1.42cm}@{\hskip2pt}@{\hskip2pt}C{1.4cm}@{\hskip2pt}@{\hskip2pt}C{1.42cm}@{\hskip6pt}c@{\hskip6pt}C{1.6cm}@{\hskip2pt}@{\hskip2pt}C{1.4cm}@{\hskip2pt}@{\hskip2pt}C{1.6cm}@{\hskip6pt}c@{\hskip6pt}C{1.6cm}@{\hskip2pt}@{\hskip2pt}C{1.8cm}@{\hskip2pt}@{\hskip2pt}C{1.8cm}@{}m{0pt}@{}}
			
			&&&\multicolumn{3}{c}{10 -- 90 Latency $t$ (ms)}&&\multicolumn{3}{c}{50 -- 50 Latency $t$ (ms)}&&\multicolumn{3}{c}{90 -- 10 Latency $t$ (ms)}& \\ \cmidrule{4-6}\cmidrule{8-10}\cmidrule{12-14}
            &&L&$t_t$&$t_p$&$t=t_t+t_p$&&$t_t$&$t_p$&$t=t_t+t_p$&&$t_t$&$t_p$&$t=t_t+t_p$& \\ \midrule[1.25pt]
			&\multirow{5}{1.20cm}{\centering No shared layers}
			&1&$14.7\pm0.3$&$89.9\pm0.0$&$104.6\pm0.3$&&
			$15.6\pm1.1$&$89.9\pm0.1$&$105.4\pm1.1$&&
			$19.4\pm5.5$&$634.4\pm417.0$&$653.8\pm413.6$& \\  
			&&2&$3.4\pm0.1$&$89.9\pm0.0$&$93.3\pm0.1$&&
			$3.8\pm0.6$&$89.9\pm0.0$&$93.6\pm0.6$&&
			$10.2\pm5.0$&$317.1\pm401.5$&$327.3\pm399.7$& \\  
			&&3&$0.9\pm0.1$&$89.9\pm0.0$&$90.7\pm0.1$&&
			$1.1\pm0.4$&$89.9\pm0.0$&$90.9\pm0.4$&&
			$4.0\pm2.9$&$198.0\pm240.0$&$201.9\pm242.4$& \\  
			&&4&$0.7\pm0.1$&$89.9\pm0.0$&$90.6\pm0.1$&&
			$0.8\pm0.3$&$89.9\pm0.0$&$90.7\pm0.3$&&
			$4.0\pm4.5$&$119.8\pm67.0$&$123.7\pm71.2$& \\  
			&&5&$0.5\pm0.1$&$89.9\pm0.0$&$90.4\pm0.1$&&
			$0.7\pm0.3$&$89.9\pm0.0$&$90.5\pm0.3$&&
			$3.0\pm2.8$&$105.1\pm34.1$&$108.1\pm36.6$& \\ \midrule[1.25pt]

			&\multirow{5}{1.20cm}{\centering First two layers shared}
			&1&$14.6\pm0.1$&$89.9\pm0.0$&$104.5\pm0.1$&&
			$15.4\pm1.2$&$89.9\pm0.1$&$105.3\pm1.2$&&
			$23.4\pm9.0$&$452.8\pm470.8$&$476.2\pm467.4$& \\  
			&&2&$3.4\pm0.1$&$89.9\pm0.0$&$93.3\pm0.1$&&
			$3.7\pm0.5$&$89.9\pm0.0$&$93.5\pm0.5$&&
			$9.6\pm8.6$&$250.6\pm389.8$&$260.2\pm388.0$& \\  
			&&3&$2.0\pm0.0$&$89.9\pm0.0$&$91.9\pm0.0$&&
			$2.2\pm0.4$&$89.9\pm0.0$&$92.1\pm0.4$&&
			$7.1\pm5.2$&$115.9\pm63.1$&$122.9\pm65.8$& \\  
			&&4&$0.8\pm0.0$&$89.9\pm0.0$&$90.7\pm0.0$&&
			$0.9\pm0.3$&$89.9\pm0.0$&$90.8\pm0.3$&&
			$2.3\pm1.6$&$90.5\pm1.3$&$92.8\pm2.2$& \\  
			&&5&$0.8\pm0.0$&$89.9\pm0.0$&$90.6\pm0.0$&&
			$0.8\pm0.2$&$89.9\pm0.0$&$90.7\pm0.2$&&
			$2.0\pm1.4$&$90.2\pm0.6$&$92.2\pm1.7$& \\ \bottomrule[1.25pt]

		\end{tabular}
	}
\subfloat[][The results with the Wi-Fi HaLow as adopted transmission technology.]{

    \begin{tabular}{@{}m{0pt}@{}C{1.1cm}@{\hskip2pt}@{\hskip2pt}C{0.4cm}@{\hskip2pt}@{\hskip2pt}C{1.42cm}@{\hskip2pt}@{\hskip2pt}C{1.4cm}@{\hskip2pt}@{\hskip2pt}C{1.42cm}@{\hskip6pt}c@{\hskip6pt}C{1.6cm}@{\hskip2pt}@{\hskip2pt}C{1.4cm}@{\hskip2pt}@{\hskip2pt}C{1.6cm}@{\hskip6pt}c@{\hskip6pt}C{1.6cm}@{\hskip2pt}@{\hskip2pt}C{1.8cm}@{\hskip2pt}@{\hskip2pt}C{1.8cm}@{}m{0pt}@{}}
        
        &&&\multicolumn{3}{c}{10 -- 90 Latency $t$ (ms)}&&\multicolumn{3}{c}{50 -- 50 Latency $t$ (ms)}&&\multicolumn{3}{c}{90 -- 10 Latency $t$ (ms)}& \\ \cmidrule{4-6}\cmidrule{8-10}\cmidrule{12-14}
        &&L&$t_t$&$t_p$&$t=t_t+t_p$&&$t_t$&$t_p$&$t=t_t+t_p$&&$t_t$&$t_p$&$t=t_t+t_p$& \\ \midrule[1.25pt]
        &\multirow{5}{1.20cm}{\centering No shared layers}
        &1&$147.2\pm2.3$&$89.9\pm0.3$&$237.2\pm2.3$&&
        $154.4\pm11.5$&$91.2\pm6.1$&$245.6\pm14.1$&&
        $187.0\pm41.4$&$624.6\pm416.9$&$811.6\pm394.1$& \\  
        &&2&$34.5\pm1.5$&$89.9\pm0.0$&$124.4\pm1.5$&&
        $37.4\pm5.3$&$90.0\pm0.6$&$127.4\pm5.4$&&
        $95.3\pm44.4$&$301.7\pm383.1$&$397.0\pm372.3$& \\  
        &&3&$8.8\pm1.2$&$89.9\pm0.3$&$98.8\pm1.3$&&
        $9.9\pm2.7$&$90.4\pm0.9$&$100.3\pm3.0$&&
        $32.9\pm28.8$&$188.1\pm228.8$&$221.0\pm254.2$& \\  
        &&4&$7.0\pm1.0$&$89.9\pm0.0$&$96.9\pm1.0$&&
        $8.1\pm2.6$&$89.9\pm0.0$&$98.0\pm2.6$&&
        $37.5\pm42.6$&$116.6\pm64.0$&$154.1\pm105.1$& \\  
        &&5&$5.3\pm0.9$&$89.9\pm0.0$&$95.2\pm0.9$&&
        $6.4\pm2.7$&$89.9\pm0.0$&$96.3\pm2.7$&&
        $28.6\pm27.0$&$103.5\pm32.6$&$132.1\pm57.2$& \\ \midrule[1.25pt]
        &\multirow{5}{1.20cm}{\centering First two layers shared}
        &1&$146.6\pm1.6$&$89.9\pm0.3$&$236.5\pm1.7$&&
        $152.3\pm9.8$&$90.7\pm1.2$&$243.1\pm10.1$&&
        $223.7\pm63.6$&$454.4\pm459.1$&$678.0\pm434.8$& \\  
        &&2&$34.1\pm0.7$&$89.9\pm0.2$&$124.0\pm0.8$&&
        $36.0\pm4.3$&$90.2\pm0.9$&$126.2\pm4.5$&&
        $85.4\pm72.2$&$252.5\pm389.8$&$338.0\pm381.7$& \\  
        &&3&$20.4\pm0.7$&$89.9\pm0.2$&$110.3\pm0.7$&&
        $21.9\pm3.9$&$90.1\pm0.6$&$112.1\pm4.0$&&
        $65.3\pm46.3$&$117.0\pm62.8$&$182.3\pm97.5$& \\  
        &&4&$8.4\pm0.3$&$89.9\pm0.2$&$98.3\pm0.4$&&
        $9.0\pm2.2$&$90.2\pm0.9$&$99.2\pm2.4$&&
        $17.9\pm14.1$&$92.1\pm1.5$&$110.0\pm14.4$& \\  
        &&5&$7.6\pm0.3$&$89.9\pm0.2$&$97.5\pm0.3$&&
        $8.2\pm2.2$&$90.0\pm0.5$&$98.2\pm2.2$&&
        $17.1\pm14.1$&$91.0\pm0.7$&$108.0\pm14.2$& \\ \bottomrule[1.25pt]
\end{tabular}
}
	\label{tab:multicnnresults}
	%
\end{table*}
This configuration encompasses a single CNN, either with or without an Early-Exit, to be placed on the considered IoT system. The methodology is tested with $L\in\{1,2,3,4\}$ and compared to the Cloud approximation ($L=5$, referred to as $L=C$), where all the computation can be placed on a single node, i.e., it can be seen as an approximation of sending data directly to the Cloud and then receiving back the result. 

The results are shown in Table~\ref{tab:singlecnnresults}, in the partition setting 50\%--50\%. Interesting results arise. First of all, the expected processing time $t_p$ is significantly reduced when the Early Exit is employed, as expected and studied in~\cite{bolukbasi2017adaptive,disabato2018reducing} for both the considered CNNs\footnote{The latency $t$ and its terms $t_t$ and $t_p$, are defined as an expected value with EX-CNNs, by weighting their values up to each layer $j$ of CNN $u$ with the probability $g_{u,j}$ of providing the classification.}. In the case of 6-layer EX-CNN, the Early-Exit allows to save about 75-84\% ($45$ to $80\%$ with \textit{Wi-Fi HaLoW}) of the latency $t$, whereas on the AlexNet 34 to 50\%.
After that, with \textit{Wi-Fi 4} transmission technology the transmission latency $t_t$ is significantly lower than the processing one $t_p$, thus it is reasonable to assume that the achieved $t_p$ is the minimum feasible in this IoT system. However, with \textit{Wi-Fi HaLow}, the minimum experimental latency ($t_p=44.93$ms for the 5-layer CNN and $t_p=1257.71$ms for the AlexNet) cannot be achieved and, in particular, the AlexNet processing with $L=1$ requires more than $3$ seconds (instead of $1.46$s with the Wi-Fi 4 transmission technology), a latency that might be unfeasible in many real applications.
The third crucial comment is about the $L=C$ case. The latency $t$ of this case and those of corresponding ones having $L\ge2$ ($L>2$ with \textit{Wi-Fi HaLow}) are almost equal (10\% increment with $L=1$)and \textit{Wi-Fi 4}), showing the capability of the proposed methodology of distributing the CNN computation among units with negligible latency increments. 
\subsubsection{Multi-CNN Configuration}
\label{subsct:resultsmulticnn}
In this configuration, two 5-layer CNNs have to be placed on the considered IoT system, with and without the first two layers shared.
Results are presented in Table~\ref{tab:multicnnresults}, for all the configurations and transmission technologies. Several comments arise. At first, in the configurations 90\%--10\%, the methodology has to often rely on STM32H7 nodes, hence the processing time is significantly increased (the computation capability $e$ of a Raspberry is 14 times greater than that of an STM32H7).
This result is even more evident when there are shared layers since the methodology can place less computation on STM32H7s.

The Wi-Fi 4 guarantees transmission latencies $t_t$s negligible w.r.t. the processing time $t_p$, that represents more than 85\% of latency $t$ (up to 96-99\% with $L\ge2$). Interestingly, the processing time is always equal to $89.9$ms, that is the experimental minimum achievable value in this IoT system. This consideration is no longer valid with the \textit{Wi-Fi HaLow}, where the two terms are comparable, especially when $L=1$. The methodology cannot indeed always achieve the minimum experimental processing latency, but sometimes it has to rely on nearby STM32H7 units, as highlighted by the non-null standard deviation of the $t_p$, in the configurations with at least 50\% of Raspberry Pi 3B+ units. Interestingly, despite the data-rate of the \textit{Wi-Fi 4} is ten times greater than that of \textit{Wi-Fi HaLow}, the latency $t$ in the harsher case with 90\% of STM32H7 units is similar for both the transmission technologies, with a maximum increment of 20\% with $L=1$ and no shared layers.

Finally, in all cases with $L \ge 2$ ($L>2$ with \textit{Wi-Fi HaLow}) the total latency $t$ is comparable to the case ($L=5$), as in Single-CNN configuration (Section~\ref{subsct:resultssinglecnn}). It is crucial to point out the importance of this result because distributing the CNN processing on various IoT units with a negligible increment in latency $t$ will allow defining a pipeline in processing  sequence of images. Indeed, when a unit has carried out the processing of CNN layers is designed to and sent the computed representation to the subsequent node, it is ready to operate on the next image, as in processor pipelines. Hence, the throughput of CNN processing can be significantly increased by processing images in a pipeline, with bottleneck the IoT unit responsible for the highest processing time.
\subsection{The Second IoT System: 50 IoT Units and Three Technological Families}
\label{subsct:harsherIoTsystem}
A second IoT system is considered with $N=50$ units belonging to three different technological families equipped with the \textit{Wi-Fi 4} transmission technology and partitioned as follows: 45\% of OrangePi Zero, 45\% of BeagleBone AI, and 10\% of Raspberry Pi 3B+. 
An example of this IoT system is shown in Figure~\ref{fig:exampleIoTsystem}, where the OrangePi Zero units are represented by a circle, the BeagleBone AI units by a square, the Raspberry Pi 3B+ units by a diamond, the sources by an asterisk, and the target unit by a circled cross.

The scenario is interesting because the most powerful IoT units in terms of both memory and computation capabilities, i.e., the Raspberry Pi 3B+, are just a few (about 5 in each simulated IoT system), whereas the remaining IoT units are characterized by contrasting peculiarities: on one hand the BeagleBone AI units have the same memory capability of the Raspberry Pi 3B+ but only 54\% of the computation one; on the other hand the OrangePi Zero have almost the same computation capability of the Raspberry Pi 3B+ (85\%, 160\% if compared to BeagleBone AI), but only a fourth of the memory capability. 
It is worth noting that both the Raspberry Pi 3B+ and BeagleBone AI can store all the layers of the considered CNNs, whereas being the OrangePi Zero units with 128 MB of RAM they cannot store all the layers of the ResNet and the (EX--)AlexNet CNNs. As a consequence, balancing between the faster OrangePi Zero units (in terms of computation capability) and the slower but with higher memory capacity BeagleBone AI units is expected in this IoT system (at least after all the Raspberry Pi 3B+ units have been considered, if enough closer).

In addition to the figures of merit presented in Section~\ref{subsct:figuresofmerit}, the number of considered nodes $\eta_x$ is taken into account, where $x$ is a technological family of IoT units, i.e. $x$ can be $R$, $O$, or $B$, representing the Raspberry Pi 3B+, the OrangePi Zero, and the BeagleBone AI units, respectively.
\subsubsection{Single-CNN Configuration}
In this configuration, 1 ResNet CNN has to be placed in this IoT system.

The results are shown in Table~\ref{tab:resultsresnet}. The processing latency $t_p$ of a ResNet CNN is extremely high but proportional to the number of operations required by such CNN to process a single image, i.e., about 7.5 times than those of the AlexNet. The methodology indeed does not encompass any optimization in processing the convolutions or any other optimization unless one re-compute the number of required operations accordingly.
Since the latency $t$ is almost only composed by the processing $t_p$ (97 to 99\% of $t$), the methodology relies only on the IoT units having the highest computation capability, as highlighted by the values of $\eta_B$ that are close to zero in almost all the cases.
\subsubsection{Multi-CNN Configuration}
\label{subsct:multicnnalexnet}
\begin{table*}[t]
    \centering
    \scriptsize
    \caption{The single CNN configuration results with $N=50$ OrangePi Zero, BeagleBone AI, and Raspberry Pi 3B+ units (with probability 45\%-45\%-10\%) and 1 ResNet CNN. The figures of merit (mean $\pm$ std) are the latency $t$, i.e., the transmission time $t_t$ plus the processing time $t_p$, and the number $\eta$ of IoT units used.}
    \begin{tabular}{@{}m{0pt}@{}C{0.4cm}C{2.25cm}C{2.25cm}C{2.25cm}cC{1.4cm}C{1.4cm}C{1.4cm}@{}m{0pt}@{}}
        &&\multicolumn{3}{c}{Latency $t$ (ms)}&&\multicolumn{3}{c}{Node usage $\eta$}&\\ \cmidrule{3-5}\cmidrule{7-10}
        &L&$t_t$&$t_p$&$t=t_t+t_p$&&$\eta_R$&$\eta_O$&$\eta_B$&\\ \midrule[1.25pt]			
        &1&$361.85\pm60.77$&$12059.62\pm430.32$&$12421.47\pm406.82$&
        &$4.92\pm1.84$&$4.05\pm1.84$&$0.03\pm0.17$&
        \\  
       &2&$249.47\pm69.22$&$11723.26\pm319.91$&$11972.72\pm313.71$&
       & $3.59\pm0.90$&$1.49\pm0.95$&$0.01\pm0.10$&
        \\  
        &3&$183.34\pm87.14$&$11645.13\pm259.90$&$11828.47\pm267.93$&
        &$2.80\pm0.55$&$0.55\pm0.86$&$0.00\pm0.06$&
        \\  
        &4&$138.99\pm85.02$&$11620.58\pm236.18$&$11759.57\pm252.58$&
        &$2.14\pm0.50$&$0.88\pm0.55$&$0.00\pm0.07$&
        \\  \cmidrule{2-5}\cmidrule{7-10}
        &C&$127.85\pm94.81$&$11602.91\pm215.07$&$11730.76\pm237.14$&
        &$0.99\pm0.11$&$0.08\pm0.34$&$0.00\pm0.00$&
        \\ \bottomrule[1.25pt]
    \end{tabular}
    \label{tab:resultsresnet}
\end{table*}
In this bunch of experiments, 1, 3, or 4 AlexNet CNNs have to be placed in the IoT system. It is worth noting that in the latter case, when $L=1$, 32 nodes out of 50 have to be used, allowing us to in-depth analyze the best placements.

The results are shown in Table~\ref{tab:multicnnresultsalexnet}, with the hyper-parameter $L$ ranging from 1 to 4 and the case $L=C$ simulating the Cloud, i.e., when all the layers of a CNN can be placed on the same IoT unit due to $L = M$\footnote{It is worth nothing that with $L=C$, the layers of an (EX--)AlexNet CNN can be placed on a single node if and only if Raspberry Pi 3B+ or BeagleBone AI IoT units are employed. If an OrangePi unit is selected by the methodology, at least another IoT unit has to be considered to place all the layers, also in this configuration.}. Several comments arise. First of all, the latency $t$ of 1 AlexNet is higher than that in the IoT system comprising only Raspberry Pi 3B+ and STM32H7 (see Section~\ref{subsct:firstiotsystem}): this is justified by the fact that in this IoT system the probability that an IoT unit is a Raspberry is 10\%, instead of 50\%, and both the OrangePi Zero and the BeagleBone AI units have a smaller computation capability.

Second, the latency $t$ of 3 AlexNet is close to that with 1 AlexNet multiplied by 3. In fact, the difference in percentage ranges from 0.6\% to 3.5\% ($L=C$ to $L=1$)\footnote{The error percentage between the latency $t_{3A}$ of placing 3 AlexNet CNNs and the latency $t_A$ of placing 1 AlexNet multiplied by 3 is computed as follows: $(t_{3A} - 3\cdot t_A) / t_{3A}$.}. When 4 AlexNet CNNs are employed, the range is slightly higher, i.e., 1 to 5\% on the same values of $L$. This result is very interesting, showing the effectiveness of the proposed methodology in placing the CNNs in the given IoT system. Moreover, by observing the values of $\eta_R$, $\eta_O$, $\eta_B$ it is clear that whenever possible the methodology relies on the fastest units, as expected by the fact that the transmission latency $t_t$ is, in all the cases, significantly smaller than the processing $t_p$ one.
Finally, as commented in Section~\ref{subsct:resultsmulticnn}, the latency $t$ with $L=1$ and $L=2$ is close to that with $L=C$, with an increment ranging from 13\% to 18\%, and from 6\% to 9\%, with 1 and 4 AlexNet CNNs, respectively, allowing us to define processing pipelines in the considered IoT system, to further reduce the latency $t$.

In Table~\ref{tab:multicnnresultsexalexnet}, the same IoT scenario is investigated with 1, 3, and 4 EX-AlexNet to be placed. The latency $t$ and its components $t_p$ and $t_t$ are defined as an expected value, by weighting the latency of each possible path within the EX-CNN by its probability. The mean numbers of nodes used $\eta_R$, $\eta_O$, and $\eta_B$ are instead computed on the longest path within the EX-CNN.

The trend in the results is analogous to the case with AlexNet CNNs, with smaller errors. The difference in percentage between placing 3 and 4 EX-AlexNet and 1 EX-AlexNet  multiplied by 3 and 4 ranges from 0.2\% to 2.5\% and from 0.4\% to 3.7\% ($L=C$ to $L=1$), respectively. Interestingly, the values of $\eta_B$ are higher in this group of experiments, showing that the methodology more often relies on closer BeagleBone AI units to place part of the EX-AlexNet computation, reasonably on the less probable path. The OrangePi Zero units are more often used in this scenario as well.
\begin{table*}[t]
	\centering
	\scriptsize
	\caption{The multi CNN configuration results with $N=50$ OrangePi Zero, BeagleBone AI, and Raspberry Pi 3B+ units (with probability 45\%-45\%-10\%) and 1 to 4 AlexNet (A) CNNs. The figures of merit (mean $\pm$ std) are the latency $t$, i.e., the transmission time $t_t$ plus the processing time $t_p$, and the number $\eta$ of IoT units used.}
	%
        \begin{tabular}{@{}m{0pt}@{}C{1.2cm}C{0.4cm}C{2cm}C{2cm}C{2cm}cC{1.4cm}C{1.4cm}C{1.4cm}@{}m{0pt}@{}}
            &&&\multicolumn{3}{c}{Latency $t$ (ms)}&&\multicolumn{3}{c}{Node usage $\eta$}&\\ \cmidrule{4-6} \cmidrule{8-10}	
      		&&L&$t_t$&$t_p$&$t=t_t+t_p$&&$\eta_R$&$\eta_O$&$\eta_B$&\\ \midrule[1.25pt]			
			&\multirow{5}{1.20cm}{\centering 1 AlexNet}
			&1&$206.22\pm26.04$&$1\,390.95\pm54.58$&$1\,597.17\pm54.33$&&
			$3.97\pm1.51$&$4.01\pm1.48$&$0.02\pm0.15$&\\  
			&&2&$151.25\pm30.93$&$1\,344.36\pm56.70$&$1\,495.61\pm60.89$&&
			$3.00\pm0.87$&$1.43\pm1.08$&$0.01\pm0.12$&\\  
			&&3&$125.55\pm33.97$&$1\,328.63\pm51.73$&$1\,454.19\pm62.16$&&
			$2.47\pm0.63$&$0.64\pm0.74$&$0.01\pm0.12$&\\  
			&&4&$119.76\pm34.24$&$1\,323.49\pm47.45$&$1\,443.25\pm59.89$&&
			$1.96\pm0.50$&$0.51\pm0.73$&$0.01\pm0.09$&\\  \cmidrule{3-6} \cmidrule{8-10}	
			&&C&$97.88\pm39.57$&$1\,311.71\pm44.03$&$1\,409.58\pm66.63$&&
			$1.00\pm0.11$&$0.17\pm0.45$&$0.01\pm0.09$&\\ \midrule[1.25pt]
            &\multirow{5}{1.20cm}{\centering 3 AlexNet}
            &1&$578.58\pm50.35$&$4\,390.39\pm104.53$&$4\,968.97\pm119.87$&&
            $5.07\pm2.01$&$17.89\pm1.72$&$1.05\pm1.24$&\\  
            &&2&$410.26\pm43.87$&$4\,211.45\pm148.26$&$4\,621.71\pm142.12$&&
            $5.05\pm1.97$&$7.72\pm1.95$&$0.11\pm0.38$&\\  
            &&3&$350.06\pm45.97$&$4\,108.70\pm154.98$&$4\,458.76\pm152.25$&&
            $4.84\pm1.71$&$4.65\pm1.98$&$0.04\pm0.26$&\\  
            &&4&$349.40\pm54.44$&$4\,035.36\pm146.02$&$4\,384.76\pm147.68$&&
            $4.46\pm1.34$&$3.13\pm1.86$&$0.04\pm0.26$&\\  \cmidrule{3-6} \cmidrule{8-10}	
            &&C&$306.24\pm76.75$&$3\,947.47\pm106.77$&$4\,253.71\pm144.39$&&
            $2.86\pm0.46$&$0.90\pm1.33$&$0.02\pm0.21$&\\ \midrule[1.25pt]
            &\multirow{5}{1.20cm}{\centering 4 AlexNet}
            &1&$772.23\pm54.88$&$5\,960.88\pm138.87$&$6\,733.11\pm156.94$&&
            $5.22\pm2.20$&$21.73\pm2.80$&$5.04\pm2.99$&\\  
            &&2&$521.34\pm36.80$&$5\,718.77\pm181.19$&$6\,240.11\pm175.43$&&
            $5.22\pm2.20$&$11.62\pm2.20$&$0.19\pm0.55$&\\  
            &&3&$437.76\pm50.07$&$5\,572.27\pm206.36$&$6\,010.03\pm196.11$&&
            $5.17\pm2.12$&$7.47\pm2.34$&$0.07\pm0.40$&\\  
            &&4&$441.12\pm59.81$&$5\,458.45\pm213.15$&$5\,899.57\pm201.50$&&
            $4.99\pm1.85$&$5.25\pm 2.36$&$0.10\pm0.38$&\\  \cmidrule{3-6} \cmidrule{8-10}	
            &&C&$401.83\pm77.75$&$5\,288.65\pm150.73$&$5\,690.48\pm183.33$&&
            $3.64\pm0.74$&$1.81\pm1.93$&$0.04\pm0.30$&\\ \midrule[1.25pt]
		\end{tabular}
	\label{tab:multicnnresultsalexnet}
\end{table*}
\begin{table*}[t]
	\centering
	\scriptsize
	\caption{The multi CNN configuration results with $N=50$ OrangePi Zero, BeagleBone AI, and Raspberry Pi 3B+ units (with probability 45\%-45\%-10\%) and 1 to 4 EX-AlexNet CNNs. The figures of merit (mean $\pm$ std) are the latency $t$, i.e., the transmission time $t_t$ plus the processing time $t_p$, and the number $\eta$ of IoT units used.}
	%
        \begin{tabular}{@{}m{0pt}@{}C{1.2cm}C{0.4cm}C{2cm}C{2cm}C{2cm}cC{1.4cm}C{1.4cm}C{1.4cm}@{}m{0pt}@{}}
            &&&\multicolumn{3}{c}{Latency $t$ (ms)}&&\multicolumn{3}{c}{Node usage $\eta$}&\\ \cmidrule{4-6} \cmidrule{8-10}	
			&&L&$t_t$&$t_p$&$t=t_t+t_p$&&$\eta_R$&$\eta_O$&$\eta_B$&\\ \midrule[1.25pt]			
			&\multirow{5}{1.20cm}{\centering 1 AlexNet}
			&1&$117.54\pm17.66$&$666.38\pm33.56$&$783.92\pm 36.27$&&
			$4.08\pm1.46$&$3.79\pm1.45$&$0.13\pm0.34$&\\  
			&&2&$93.87\pm24.95$&$643.84\pm34.46$&$737.71\pm40.10$&&
			$3.09\pm0.82$&$1.51\pm1.00$&$0.07\pm0.25$&\\  
			&&3&$77.03\pm25.44$&$641.06\pm34.80$&$718.09\pm41.04$&&
			$2.66\pm0.58$&$0.44\pm0.67$&$0.02\pm0.13$&\\  
			&&4&$74.56\pm24.62$&$640.61\pm33.94$&$715.17\pm40.58$&&
			$1.98\pm0.49$&$0.44\pm0.71$&$0.00\pm0.06$&\\  \cmidrule{3-6} \cmidrule{8-10}	
			&&C&$68.15\pm24.14$&$639.05\pm 34.41$&$707.20\pm42.28$&&
			$1.00\pm0.06$&$0.24\pm0.49$&$0.01\pm0.08$&\\ \midrule[1.25pt]
            &\multirow{5}{1.20cm}{\centering 3 AlexNet}
            &1&$335.42\pm21.31$&$2\,078.03\pm54.60$&$2\,413.45\pm59.52$&&
            $4.95\pm2.08$&$17.26\pm1.72$&$1.79\pm1.29$&\\  
            &&2&$268.92\pm38.28$&$1989.59\pm72.05$&$2258.52\pm72.76$&&
            $4.91\pm2.03$&$8.36\pm1.82$&$0.55\pm0.78$&\\  
            &&3&$222.56\pm39.05$&$1971.77\pm75.12$&$2194.33\pm75.51$&&
            $4.79\pm1.90$&$4.84\pm2.22$&$0.17\pm0.59$&\\  
            &&4&$222.98\pm39.87$&$1946.47\pm73.10$&$2169.45\pm76.50$&&
            $4.37\pm1.43$&$3.37\pm2.00$&$0.11\pm0.36$&\\  \cmidrule{3-6} \cmidrule{8-10}	
            &&C&$211.27\pm42.69$&$1914.79\pm64.03$&$ 2126.06\pm75.80$&&
            $2.87\pm0.51$&$1.08\pm1.36$&$0.03\pm0.17$&\\ \midrule[1.25pt]
            &\multirow{5}{1.20cm}{\centering 4 AlexNet}
            &1&$452.33\pm29.81$&$2803.39\pm67.85$&$3255.73\pm78.62$&&
            $4.95\pm2.17$&$21.52\pm2.72$&$5.53\pm2.56$&\\  
            &&2&$346.37\pm43.73$&$2689.37\pm94.61$&$3035.74\pm93.17$&&
            $4.95\pm2.17$&$12.39\pm1.79$&$1.19\pm1.35$&\\  
            &&3&$284.80\pm46.16$&$2666.04\pm100.07$&$2950.84\pm96.52$&&
            $4.93\pm2.13$&$8.07\pm2.15$&$0.72\pm1.05$&\\  
            &&4&$289.94\pm44.66$&$2619.44\pm103.84$&$2909.38\pm102.62$&&
            $4.79\pm1.89$&$5.59\pm 2.24$&$0.25\pm0.68$&\\  \cmidrule{3-6} \cmidrule{8-10}	
            &&C&$281.67\pm50.74$&$2558.70\pm98.07$&$2840.36\pm105.54$&&
            $3.63\pm0.95$&$1.86\pm2.12$&$0.12\pm0.51$&\\ \midrule[1.25pt]
		\end{tabular}
	\label{tab:multicnnresultsexalexnet}
\end{table*}
\input{figure_example.tex}
%
%
%
\section{Porting a CNN to a real IoT System}
\label{subsct:realImplementation}
The methodology has been also applied to the placement of the 5-layer CNN described in Section \ref{subsct:singleCNN} and depicted in Fig. 1(a) on a real technological scenario comprising two \textit{STM32H7}s and one \textit{Raspberry Pi 3B+}. The transmission technology is the \textit{Wi-Fi 4} and the connectivity is provided locally by the GL.iNet GL-MT300N-V2 router. 
The goal of this experiment is to compare the figures of merit $t$, $t_t$, and $t_p$ of the CNN placement provided by the methodology with those of the real CNN placement in the considered technological scenario.
With $L=4$ and the nodes equally spaced (each at distance 1 from each other), the methodology assigned the first four layers of the CNN to the Raspberry and the fifth layer to one of the two STM32H7s. The measured transmission and processing times, shown in Table~\ref{tab:benchmark}, are particularly interesting, showing that the experimental transmission time $t_t$ is almost equal to the methodology estimation, whereas the experimental processing time $t_p$ is 30\% larger. This is justified by the fact the model considered only the multiplications. In particular, the first four layers on Raspberry Pi 3B+ took 68.29 ms instead of 44.93 ms, whereas the fifth layer on STM32H7 176 $\mu$s (32 $\mu$s with code optimization) instead of 50 $\mu$s.

It is worth nothing that the measured processing time $t_p$ of the AlexNet on the Raspberry Pi 3B+ (median over 100 runs) is 1119.47 ms, whereas the one provided by the methodology is 1257.71 ms, showing that the approximation given in Section~\ref{sct:theMethodology} well describe this technological scenario.
\section{Conclusions}
\label{sct:conclusions}
The aim of this paper was to introduce a novel effective methodology for the optimal placement of CNNs on IoT systems. Such a methodology is general enough to be applied to multiple sources of data and multiple CNNs operating in the same IoT system and has been formalized as an optimization problem where the latency between image acquisitions and the decision making is minimized.
Future works will encompass the extension of the methodology with dynamic scheduling and routing algorithms for IoT to deal with communications failures and mobile units as well as improve the performance by exploiting the resources that are made available by early-exit mechanisms. 
\begin{table}[t]
%
	\centering
	\scriptsize
	\caption{Experimental benchmark results with equally spaced nodes and a 5-layer CNN. 
    The figure of merit is the latency $t$ (transmission $t_t$ plus processing $t_p$).}
	%

	\begin{tabular}{@{}m{0pt}@{}C{0.4cm}C{1.5cm}C{1cm}C{1cm}C{1.5cm}@{}m{0pt}@{}}
		&L&Case&\multicolumn{3}{c}{Wi-Fi 4 Latency (ms)}& \\ \cmidrule{4-7}
		&&&$t_t$&$t_p$&$t=t_t+t_p$& \\ \midrule[1.25pt]
		&\multirow{2}{*}{\centering 4}&Model&0.37&44.98&45.35&\\ \cmidrule{3-7}
		&&Experimental&0.42&68.47&68.89&\\\bottomrule[1.25pt]
	\end{tabular}
	\label{tab:benchmark}
    \vspace*{-0.25cm}
	%
%
\end{table}

%
%

\ifCLASSOPTIONcaptionsoff
  \newpage
\fi



%
%
%
\bibliography{bibliography}
\bibliographystyle{IEEEtran}

\end{document}